\documentclass{article}
\usepackage[OT1]{fontenc}
\usepackage{iclr2027_preprint,times}

\usepackage{amsmath,amssymb}
\IfFileExists{math_commands.tex}{%%%%% NEW MATH DEFINITIONS %%%%%

\usepackage{amsmath,amsfonts,bm}

\def\eqref#1{equation~\ref{#1}}
\def\1{\bm{1}}

\DeclareMathAlphabet{\mathsfit}{\encodingdefault}{\sfdefault}{m}{sl}
\SetMathAlphabet{\mathsfit}{bold}{\encodingdefault}{\sfdefault}{bx}{n}

}{}

\usepackage{hyperref}
\hypersetup{hidelinks}
\usepackage{url}
\usepackage{booktabs,graphicx}
\usepackage{adjustbox,multirow}
\usepackage{tikz}
\definecolor{fpdrop}{RGB}{160,55,55}

\title{Beyond Reconstruction Loss in Post-Training Quantization: Balanced Fitting for Large Vision-Language Models}

\author{
\textbf{Minchan Kang}\textsuperscript{1}, \textbf{Kyeonghye Park}\textsuperscript{1}, \textbf{Seungyeon Sa}\textsuperscript{1}, \textbf{Seoyoung Cho}\textsuperscript{1}, \textbf{Daeshik Kim}\textsuperscript{1}\thanks{Equal correspondence: Daeshik Kim and Yucheol Cho.}, \textbf{Yucheol Cho}\textsuperscript{2}\footnotemark[1] \\
\textsuperscript{1}Korea Advanced Institute of Science and Technology (KAIST), \textsuperscript{2}Hanbat National University \\
{\small\texttt{\{mc.kang,pkhpjhs,lucy.sa,52tjdud,daeshik\}@kaist.ac.kr; yccho@hanbat.ac.kr}}
}

\usetikzlibrary{shapes.geometric}
\begin{document}

\maketitle
\begin{abstract}
Post-training quantization (PTQ) enables efficient deployment of large vision-language models (LVLMs), but is typically calibrated on a small set while expected to generalize across diverse downstream tasks. Although recent PTQ methods for LVLMs incorporate sensitivity signals, they still minimize reconstruction loss with respect to the full-precision model, potentially over-preserving FP behavior and calibration-specific bias. Rather than treating quantization solely as an error to be minimized, we observe that it can also provide beneficial regularization for certain layers and modalities. Motivated by this observation, we propose Balanced Fitting, a quantization effect-based framework that balances precision and regularization beyond reconstruction-based optimization. By measuring layer- and component-wise quantization effects for weights, vision activations, and text activations, Balanced Fitting combines fine-grained fitting for sensitive components with coarser fitting to exploit potential regularization benefits. Experiments on multiple LVLMs show that our method consistently outperforms prior PTQ approaches under both weight-only and weight-activation quantization, while lower reconstruction loss does not reliably translate into better downstream performance.
The source code is publicly available at \url{https://github.com/kmc3661/BFQ}.

\end{abstract}

\section{Introduction}

Large Vision-Language Models (LVLMs) have achieved strong multimodal reasoning performance across diverse tasks~\citep{liu2023visual}, but at the cost of high computational and memory overhead. This has motivated research on model compression, including token pruning~\citep{rao2021dynamicvit,bolya2022token}, knowledge distillation~\citep{hinton2015distilling,li2023distilling}, and quantization~\citep{jacob2018quantization,zhou2016dorefa}. Among these, post-training quantization (PTQ) is particularly practical, as it reduces memory and computation using low-precision representations and enables efficient compression with only a small calibration set~\citep{nagel2020up}.

\begin{figure}[t]
    \centering
    \includegraphics[width=\linewidth]{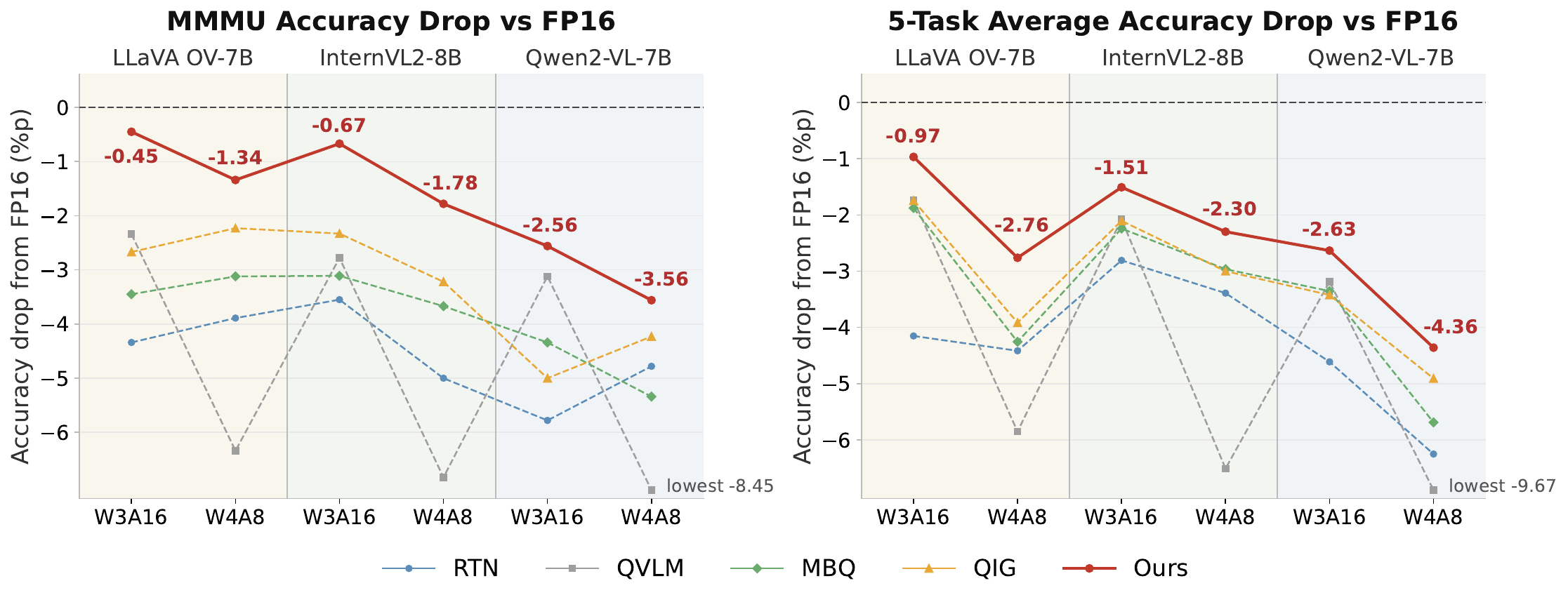}
    \caption{Accuracy drop from the full-precision (FP16) model across multiple LVLMs under low-bit settings (W3A16 and W4A8). Results are reported on the MMMU benchmark (left) and the average across five multimodal benchmarks (right). Our method consistently achieves the lowest accuracy drop across all models and configurations among state-of-the-art PTQ methods for LVLMs.}
    \label{fig:performance_comparison}
\end{figure}

While PTQ has shown strong performance in large language models (LLMs)~\citep{gptq, awq, smoothquant}, extending it to LVLMs remains challenging due to heterogeneous multimodal representations. Recent LVLM-specific methods incorporate sensitivity signals, including cross-layer dependencies (Q-VLM~\citep{qvlm}), modality-wise gradients (MBQ~\citep{mbq}), and token-level importance (QIG~\citep{qig}). Despite these advances, they share a common objective of optimizing quantization by minimizing reconstruction error with respect to the full-precision (FP) model on a small calibration set.

We argue that this paradigm is fundamentally limited. First, prior work~\citep{chen2021quantization,liang2021pruning,kuzmin2023pruning,wang2020differentiable} has reported that compression can sometimes improve performance, suggesting that the full-precision (FP) model may contain overfitted or suboptimal components. This suggests that quantization is not always a harmful perturbation; in some layers or components, it may act as an implicit regularizer. We further observe similar phenomena in LVLMs, where different components (weights, vision tokens, and text tokens) exhibit varying sensitivity across layers, and some benefit from quantization (Fig.~\ref{fig:motivation}(a)). Second, the calibration set is inherently limited and may not reflect the full downstream distribution, introducing bias. Consequently, minimizing reconstruction error with respect to the FP model on a limited calibration set can preserve overfitted FP behavior and calibration-set bias, leading to suboptimal generalization.

To address these limitations, we propose Balanced Fitting, a novel PTQ framework for LVLMs that goes beyond local reconstruction error minimization. As illustrated in Figure~\ref{fig:motivation}(b), our key idea is to adaptively control the fitting strength of quantization using component-wise effects, balancing precise fitting for sensitive components and coarser fitting where quantization serves as regularization. Specifically, we first measure the quantization effect of weights, vision activations, and text activations at each layer, indicating whether quantization disrupts important information or provides beneficial regularization on calibration samples. Based on these effects, we adapt fitting capacity using component-aware scores, encouraging finer-grained fitting for sensitive components and coarser fitting where quantization serves as regularization. This leads to a balanced distribution of quantization resources that better preserves generalization.

Extensive experiments on state-of-the-art LVLMs demonstrate the effectiveness of our approach. As shown in Figure~\ref{fig:performance_comparison}, our method consistently outperforms prior PTQ methods for LVLMs under both weight-only and weight-activation settings (W3A16 and W4A8), substantially reducing the accuracy drop from the full-precision model. Notably, on MMMU, a challenging benchmark requiring broad multidisciplinary knowledge and visual reasoning, our approach achieves a 2.43× average and up to 5.2× reduction in accuracy drop relative to the strongest prior PTQ baseline across models and bit-widths, demonstrating robust generalization on complex multimodal tasks.

\section{Related Work}
\subsection{Large Vision Language Models}
Large language models (LLMs) have demonstrated strong reasoning capabilities across diverse tasks~\citep{brown2020language, chowdhery2023palm, touvron2023llama}. Large vision-language models (LVLMs) project features from vision encoders such as ViT~\citep{dosovitskiy2020image} and CLIP~\citep{clip} into the language embedding space and combine them with text tokens for multimodal reasoning~\citep{alayrac2022flamingo,li2023blip}. Representative models include LLaVA-OneVision~\citep{llavaov}, InternVL~\citep{internvl}, and Qwen-VL~\citep{qwen2}. While prior work primarily focuses on improving cross-modal alignment and reasoning, we instead focus on efficient deployment, particularly post-training quantization for LVLMs.

\subsection{Post-Training Quantization}

Post-training quantization (PTQ) is a widely used compression technique that reduces model precision without retraining~\citep{banner2019post}. For vision transformers, prior work addresses challenges in token-wise representations and attention through various calibration and quantization strategies~\citep{liu2021post,yuan2022ptq4vit,li2023vit,lin2021fq}. For large language models (LLMs), existing approaches mitigate quantization errors caused by activation outliers and model scale~\citep{yao2022zeroquant, wei2022outlier, liu2023noisyquant}. PTQ has also been extended to large vision-language models (LVLMs). Representative methods incorporate sensitivity signals such as cross-layer dependencies (Q-VLM~\citep{qvlm}), modality-wise gradients (MBQ~\citep{mbq}), and token-level importance (QIG~\citep{qig}), yet still rely on reconstruction-based optimization over limited calibration data. Our approach instead adapts fitting strength to component-wise quantization effects, preserving sensitive components while allowing beneficial deviations from the FP model.

\begin{figure}[t]
    \centering
    \includegraphics[width=\linewidth]{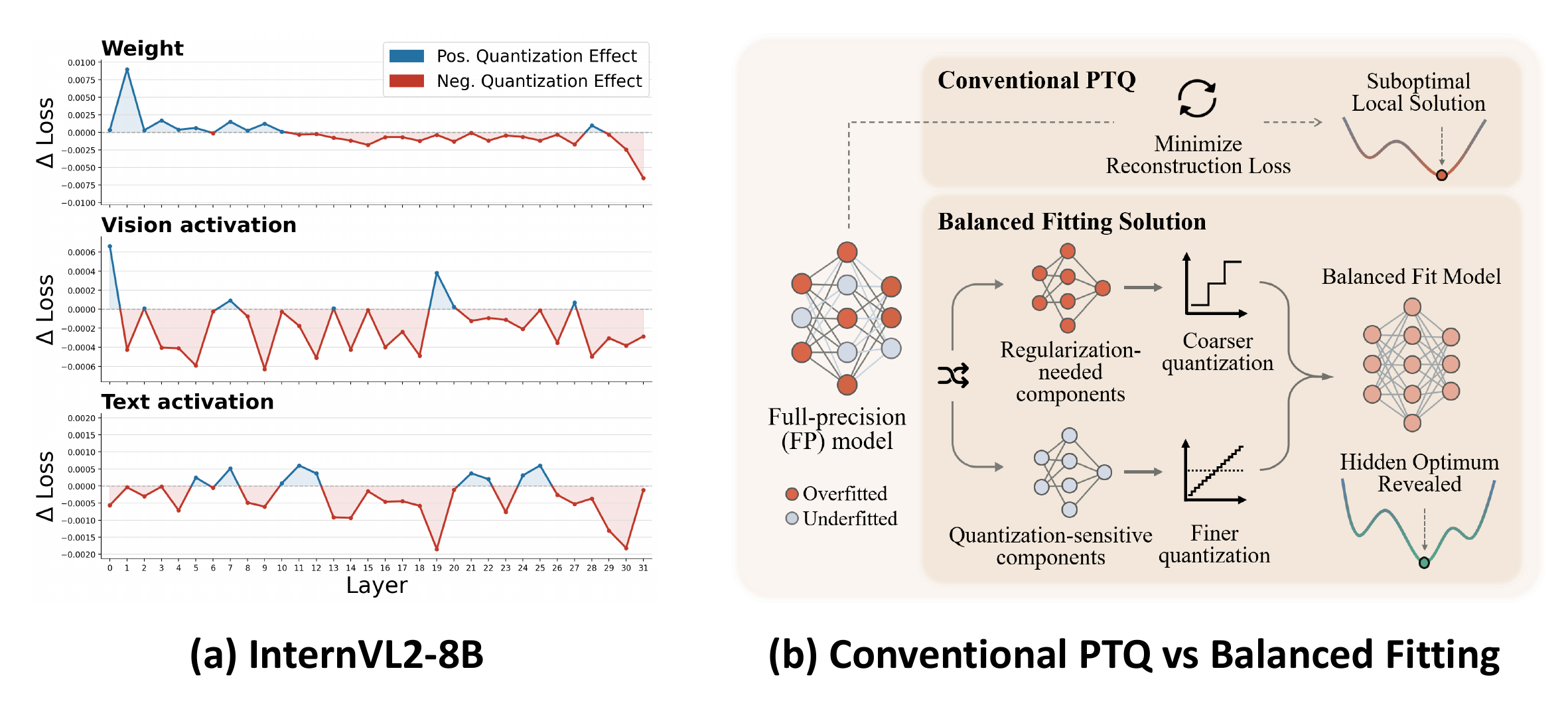}
    \caption{Motivation of our approach. (a) Component-wise quantization effects across layers reveal heterogeneous sensitivity across weights and modalities, where some components benefit from quantization while others are sensitive. (b) Conventional PTQ uniformly minimizes reconstruction error, leading to suboptimal solutions, whereas our Balanced Fitting adapts fitting strength using component-wise effects, improving generalization.}
    \label{fig:motivation}
\end{figure}

\section{Method}
\subsection{Preliminaries}

Quantization maps full-precision weights and activations to a discrete set of low-bit values, reducing memory and computation. We use uniform integer quantization, with group-wise affine quantization for weights and symmetric per-token quantization for activations. The format $\mathrm{W}x\mathrm{A}y$ denotes $x$-bit weights and $y$-bit activations; for example, W4A8 uses 4-bit weights and 8-bit activations.

Let $\mathbf{W}$ and $\mathbf{X}$ denote the full-precision weight and activation matrices, respectively. PTQ commonly minimizes the reconstruction error between quantized and full-precision outputs. Existing LVLM PTQ methods~\citep{qvlm,mbq,qig} use channel-wise equalization (CWE) to mitigate activation outliers by inversely rescaling $\mathbf{W}$ and $\mathbf{X}$. The channel-wise scaling vector $\mathbf{E}$ is optimized to minimize reconstruction error:
\begin{equation}
\mathbf{E}^* = \arg\min_{\mathbf{E}} 
\left\|
Q(\mathbf{W} \odot \mathbf{E}) \, 
Q(\mathbf{E}^{-1} \odot \mathbf{X}) 
- \mathbf{W}\mathbf{X}
\right\|_2^2,
\end{equation}
where $Q(\cdot)$ denotes quantization and $\odot$ denotes channel-wise scaling. We retain the quantization operator and reconstruction objective, changing only the number of CWE search candidates per layer.

\begin{figure}[t]
    \centering
    \includegraphics[width=\linewidth]{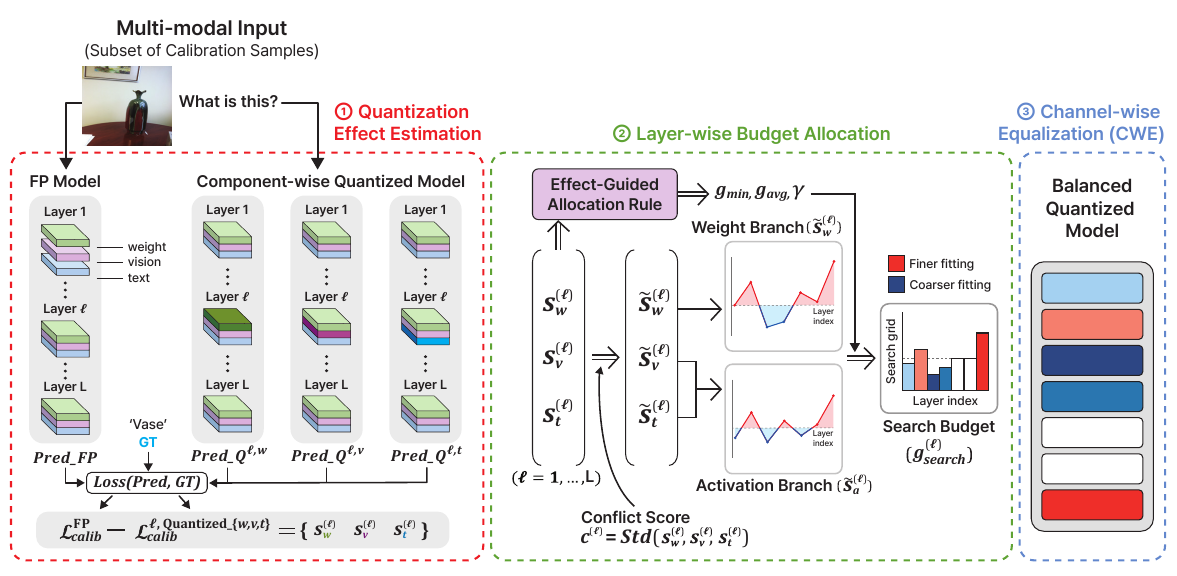}
    \caption{
Overview of the proposed balanced fitting framework. 
We first estimate layer- and component-wise quantization effects by selectively quantizing weights and activations. 
These effects are then used to guide a layer-wise budget allocation policy that adaptively controls the search granularity. 
The resulting allocation balances precise fitting for quantization-sensitive components and coarser fitting for others, improving generalization.
}
    \label{fig:overall}
\end{figure}

\subsection{Layer- and Component-wise Quantization Effects}
As illustrated in Figure~\ref{fig:overall}, we first estimate layer- and component-wise quantization effects by selectively quantizing each component and measuring its impact on the calibration loss. Instead of assigning uniform fitting strength across layers, our method characterizes how each component responds to quantization and uses this information to guide quantization budget allocation. Let $\ell \in \{1,\dots,L\}$ denote the transformer layer index. 
For each layer, we consider quantizing one component at a time with the same scheme used in the target PTQ setting, where $c \in \{w,v,t\}$ indicates weights, vision activations, and text activations, respectively. Given a calibration set, we define the calibration loss on a small subset as the token-level negative log-likelihood (NLL) of ground-truth target tokens:

\begin{equation}
\mathcal{L}_{\mathrm{calib}}
=
\frac{1}{N}
\sum_{i=1}^{N}
\frac{1}{T_i}
\sum_{r=1}^{T_i}
-\log p_{\theta}\!\left(y_{i,r}\mid x_i, y_{i,<r}\right),
\end{equation}
where $N$ is the calibration-subset size, $x_i$ the $i$-th multimodal input, $T_i$ the number of valid target tokens, and $y_{i,r}$ the $r$-th ground-truth token. Here, $p_\theta$ is the model output distribution under the current quantization setting. We then quantize only component $c$ at layer $\ell$ while keeping all others unchanged and define the quantization effect as

\begin{equation}
\label{eq:quant_effect}
s_c^{(\ell)}
=
\mathcal{L}_{\mathrm{calib}}^{(\mathrm{FP})}
-
\mathcal{L}_{\mathrm{calib}}^{(\ell,c\text{-quantized})},
\quad c \in \{w,v,t\}.
\end{equation}
A positive value of $s_c^{(\ell)}$ indicates that quantization reduces the calibration loss, suggesting a potential regularization benefit for that component. Conversely, a negative value indicates that quantization increases the loss, implying that preserving full-precision information is important. Figure~\ref{fig:motivation}(a) illustrates the measured $s_c^{(\ell)}$ on the InternVL2-8B model~\citep{internvl}. The results reveal that the impact of quantization varies significantly across layers and components (weights, vision activations, and text activations), highlighting the need for quantization adapted to this heterogeneity.

Based on this observation, we transform $s_c^{(\ell)}$ into an importance score for budget allocation. 
Sensitive components receive positive importance scores, whereas components that benefit from quantization receive negative scores to encourage coarser fitting:
\begin{equation}
\label{eq:importance_score}
\hat{s}_c^{(\ell)} =
\begin{cases}
-\alpha\, s_c^{(\ell)}, & s_c^{(\ell)} \ge 0, \\
-s_c^{(\ell)}, & s_c^{(\ell)} < 0,
\end{cases}
\end{equation}
The transformed score $\hat{s}_c^{(\ell)}$ represents the resulting importance of component $c$ at layer $\ell$, where $\alpha \in [0,1]$ controls how strongly positive quantization effects are down-weighted in the importance score. For activations, we aggregate the visual and textual components to obtain a unified importance score, 
$\hat{s}_a^{(\ell)} = \hat{s}_v^{(\ell)} + \hat{s}_t^{(\ell)}$, 
where $\hat{s}_a^{(\ell)}$ denotes the activation importance score at layer $\ell$. 

In practice, different components within the same layer can exhibit heterogeneous responses to quantization. 
When such disagreement occurs, directly aggregating component-wise quantization effects may obscure vulnerable components, potentially causing sensitive components to receive insufficient search capacity. To address this issue, we introduce a layer-wise conflict score and incorporate it into the final importance scores as follows:
\begin{equation}
\label{eq:conflict_score}
\begin{aligned}
\kappa^{(\ell)} &= \mathrm{Std}\!\left(s_w^{(\ell)}, s_v^{(\ell)}, s_t^{(\ell)}\right), \\
\tilde{s}_w^{(\ell)} &= \hat{s}_w^{(\ell)} + \beta\,\kappa^{(\ell)}, \quad
\tilde{s}_a^{(\ell)} = \hat{s}_a^{(\ell)} + \beta\,\kappa^{(\ell)}.
\end{aligned}
\end{equation}
where $\kappa^{(\ell)}$ measures the degree of disagreement among component-wise quantization effects in the $\ell$-th layer, and $\beta$ controls how strongly this disagreement is reflected in the final importance scores, thereby encouraging additional allocation to layers with heterogeneous sensitivity. This dispersion captures differences in magnitude as well as sign. In all experiments, we set $\alpha = 0.5$ and $\beta = 0.5$.

\subsection{Layer-wise Budget Allocation}

We allocate the search budget across layers based on the importance scores computed in the previous section. 
Concretely, each layer $\ell$ is assigned a discrete budget $g^{(\ell)}$, representing the number of grid points used to optimize channel-wise equalization (CWE). We define a budget policy with three quantities: the minimum per-layer budget $g_{\min}$, the target average budget $g_{\mathrm{avg}}$, and the concentration factor $\gamma$. These values are automatically determined from the distribution of quantization effects $\{s_c^{(\ell)}\}$ using our Quantization Effect-Guided Allocation Rule, with details provided in Appendix~\ref{sec:auto_rule}. Each layer is first assigned the minimum budget $g_{\min}$, and the total budget is defined as
\begin{equation}
\begin{aligned}
G_{\min} &= L \cdot g_{\min}, \quad 
G_{\mathrm{target}} = \mathrm{round}(L \cdot g_{\mathrm{avg}}), \quad
R = G_{\mathrm{target}} - G_{\min}.
\end{aligned}
\end{equation}
Here, $G_{\min}$ is the minimum total budget obtained by assigning $g_{\min}$ to all $L$ layers, $G_{\mathrm{target}}$ is the desired total budget determined by the target average budget $g_{\mathrm{avg}}$, and $R$ denotes the remaining budget to be distributed across layers. Let $u^{(\ell)}$ denote the branch-specific importance score, where $u^{(\ell)}=\tilde{s}_w^{(\ell)}$ for the weight branch and $u^{(\ell)}=\tilde{s}_a^{(\ell)}$ for the activation branch. We then transform these scores into allocation weights and distribute the remaining budget accordingly:
\begin{equation}
\begin{aligned}
t^{(\ell)} &= \max\!\left(u^{(\ell)}-q_{\mathrm{low}},0\right),
\qquad
p^{(\ell)} = \frac{\left(t^{(\ell)}\right)^{\gamma}}
{\sum_{j=1}^{L}\left(t^{(j)}\right)^{\gamma}}, \\
g^{(\ell)} &= g_{\min} + \mathrm{round}(R \cdot p^{(\ell)}).
\end{aligned}
\end{equation}
where $q_{\mathrm{low}}$ is the 50th percentile of the branch-wise score distribution and $\gamma$ controls how strongly the remaining budget is concentrated on high-importance layers. Layers below this quantile remain at $g_{\min}$, while upper-tail layers receive additional budget according to $p^{(\ell)}$; rounding is adjusted to preserve $G_{\mathrm{target}}$. Thus, smaller budgets yield coarser CWE search that can act as regularization, whereas additional budget enables more precise fitting.

Although importance scores are computed at the component level, we allocate the search budget at the layer level because CWE parameters are jointly optimized within each layer and component effects are inherently coupled. Using a unified budget per layer enables efficient exploration of candidate configurations that account for these interactions. Therefore, we allocate a single search budget $g_{\mathrm{search}}^{(\ell)}$ per layer, determined by the quantization setting.

For W3A16, only the weight branch is used, and the search budget is $g_{\mathrm{search}}^{(\ell)} = g_w^{(\ell)}.$
For W4A8, both the weight and activation branches contribute to the final search budget. 
We first compute the total positive importance of each branch and derive global branch weights:

\begin{equation}
\begin{aligned}
S_w &= \sum_{\ell} \max(\tilde{s}_w^{(\ell)}, 0), \qquad
S_a = \sum_{\ell} \max(\tilde{s}_a^{(\ell)}, 0), \\
\rho_w &= \frac{S_w}{S_w + S_a}, \quad \rho_a = 1-\rho_w, \qquad
g_{\mathrm{search}}^{(\ell)} =
\mathrm{round}\!\left(\rho_w g_w^{(\ell)} + \rho_a g_a^{(\ell)}\right).
\end{aligned}
\end{equation}

Here, $S_w$ and $S_a$ denote the total positive importance mass of the weight and activation branches, respectively. They summarize how strongly each branch demands additional search capacity across all layers. The normalized branch weights $\rho_w$ and $\rho_a$ determine each branch’s contribution to the final budget. The branch-wise budgets $g_w^{(\ell)}$ and $g_a^{(\ell)}$ follow the allocation rule above; their rounded weighted combination gives the W4A8 search budget $g_{\mathrm{search}}^{(\ell)}$.

\subsection{CWE with budget-controlled search}
Given the allocated layer-wise search budget $g_{\mathrm{search}}^{(\ell)}$, we apply channel-wise equalization (CWE) to optimize the scaling factors for each layer.
For each linear layer, we search for an optimal scaling vector $E$ within a discrete set of candidate values determined by the allocated budget. The optimization follows the standard CWE objective:
\begin{equation}
E^* = \arg\min_E \sum_{i=1}^{M} 
\left\|
Q_W(W \odot E)\, Q_X(E^{-1} \odot X_i) - W X_i
\right\|_2^2,
\end{equation}
where $X_i$ denotes the activation of the $i$-th calibration token; activation quantization is omitted for weight-only settings. Unlike prior approaches that incorporate sensitivity signals while still applying uniform fitting strength across layers, we preserve the original CWE formulation and instead control the granularity of the scale search via the allocated budget. Larger budgets enable finer-grained search for quantization-sensitive layers, while smaller budgets enforce coarser search, potentially reducing overfitting. This design aims to improve generalization by controlling the fitting granularity without modifying the underlying objective.

\begin{table}[!t]
\centering
\caption{Quantitative results across three LVLMs and five benchmarks. $\Delta$FP: Avg. $-$ FP Avg. (pp).}
\label{tab:main_results}
\small
\setlength{\tabcolsep}{3pt}
\begin{tabular}{lllccccccr}
\toprule
Model & Bitwidth & Method & MMMU & VizWiz & ScienceQA & ChartQA & AI2D & Avg. & $\Delta$FP \\
\midrule

\multirow{11}{*}{LLaVA-OV-7B}
& FP16 & - & 46.56 & 58.71 & 95.84 & 80.00 & 81.28 & 72.48 & --- \\
\cmidrule{2-10}
& \multirow{5}{*}{W3A16}
& RTN  & 42.22 & 57.00 & 94.55 & 68.88 & 78.98 & 68.33 & \textcolor{fpdrop}{$-4.15$} \\
& & Q-VLM & 44.22 & 59.72 & 94.40 & 76.76 & 78.63 & 70.75 & \textcolor{fpdrop}{$-1.73$} \\
& & MBQ  & 43.11 & 59.88 & 94.74 & 77.00 & 78.27 & 70.60 & \textcolor{fpdrop}{$-1.88$} \\
& & QIG  & 43.89 & 59.54 & 94.65 & 77.12 & 78.47 & 70.73 & \textcolor{fpdrop}{$-1.75$} \\
& & Ours & \textbf{46.11} & \textbf{60.28} & \textbf{94.94} & \textbf{77.20} & \textbf{79.02} & \textbf{71.51} & \textcolor{fpdrop}{$\mathbf{-0.97}$} \\
\cmidrule{2-10}
& \multirow{5}{*}{W4A8}
& RTN  & 42.67 & 54.92 & 94.10 & 70.28 & 78.34 & 68.06 & \textcolor{fpdrop}{$-4.42$} \\
& & Q-VLM & 40.22 & 51.93 & 92.36 & 71.48 & 77.14 & 66.63 & \textcolor{fpdrop}{$-5.85$} \\
& & MBQ  & 43.44 & 55.00 & \textbf{94.25} & 70.44 & 77.98 & 68.22 & \textcolor{fpdrop}{$-4.26$} \\
& & QIG  & 44.33 & 52.48 & 93.70 & 75.00 & 77.30 & 68.56 & \textcolor{fpdrop}{$-3.92$} \\
& & Ours & \textbf{45.22} & \textbf{55.78} & 94.00 & \textbf{75.08} & \textbf{78.50} & \textbf{69.72} & \textcolor{fpdrop}{$\mathbf{-2.76}$} \\
\midrule

\multirow{11}{*}{InternVL2-8B}
& FP16 & - & 48.11 & 60.67 & 97.07 & 82.48 & 82.35 & 74.14 & --- \\
\cmidrule{2-10}
& \multirow{5}{*}{W3A16}
& RTN  & 44.56 & 55.74 & \textbf{96.28} & 79.52 & \textbf{80.54} & 71.33 & \textcolor{fpdrop}{$-2.81$} \\
& & Q-VLM & 45.33 & 59.60 & 96.18 & 79.00 & 80.18 & 72.06 & \textcolor{fpdrop}{$-2.08$} \\
& & MBQ  & 45.00 & \textbf{59.87} & 96.18 & 78.76 & 79.66 & 71.89 & \textcolor{fpdrop}{$-2.25$} \\
& & QIG  & 45.78 & 58.42 & 96.13 & 79.52 & 80.31 & 72.03 & \textcolor{fpdrop}{$-2.11$} \\
& & Ours & \textbf{47.44} & 59.77 & 96.18 & \textbf{80.12} & 79.63 & \textbf{72.63} & \textcolor{fpdrop}{$\mathbf{-1.51}$} \\
\cmidrule{2-10}
& \multirow{5}{*}{W4A8}
& RTN  & 43.11 & 57.04 & 96.18 & 78.48 & 78.92 & 70.75 & \textcolor{fpdrop}{$-3.39$} \\
& & Q-VLM & 39.89 & 56.57 & 93.26 & 73.60 & 74.81 & 67.63 & \textcolor{fpdrop}{$-6.51$} \\
& & MBQ  & 44.44 & 57.59 & 96.33 & 78.16 & 79.34 & 71.17 & \textcolor{fpdrop}{$-2.97$} \\
& & QIG  & 44.89 & 55.72 & \textbf{96.58} & \textbf{78.76} & 79.73 & 71.14 & \textcolor{fpdrop}{$-3.00$} \\
& & Ours & \textbf{46.33} & \textbf{58.01} & \textbf{96.58} & 78.44 & \textbf{79.83} & \textbf{71.84} & \textcolor{fpdrop}{$\mathbf{-2.30}$} \\
\midrule

\multirow{11}{*}{Qwen2-VL-7B}
& FP16 & - & 50.56 & 68.91 & 84.88 & 81.56 & 80.08 & 73.20 & --- \\
\cmidrule{2-10}
& \multirow{5}{*}{W3A16}
& RTN  & 44.78 & 66.28 & 81.41 & 73.84 & 76.62 & 68.59 & \textcolor{fpdrop}{$-4.61$} \\
& & Q-VLM & 47.44 & 66.73 & 79.97 & \textbf{78.36} & 77.56 & 70.01 & \textcolor{fpdrop}{$-3.19$} \\
& & MBQ  & 46.22 & 64.87 & 82.80 & 78.00 & 77.33 & 69.84 & \textcolor{fpdrop}{$-3.36$} \\
& & QIG  & 45.56 & 64.95 & \textbf{82.90} & 78.16 & 77.30 & 69.77 & \textcolor{fpdrop}{$-3.43$} \\
& & Ours & \textbf{48.00} & \textbf{66.86} & 82.30 & 77.84 & \textbf{77.82} & \textbf{70.56} & \textcolor{fpdrop}{$\mathbf{-2.64}$} \\
\cmidrule{2-10}
& \multirow{5}{*}{W4A8}
& RTN  & 45.78 & 58.59 & 79.23 & 74.96 & 76.17 & 66.95 & \textcolor{fpdrop}{$-6.25$} \\
& & Q-VLM & 42.11 & 51.30 & 80.52 & 70.72 & 72.99 & 63.53 & \textcolor{fpdrop}{$-9.67$} \\
& & MBQ  & 45.22 & 58.99 & 79.42 & 77.00 & 76.91 & 67.51 & \textcolor{fpdrop}{$-5.69$} \\
& & QIG  & 46.33 & 61.26 & 79.57 & 76.76 & \textbf{77.53} & 68.29 & \textcolor{fpdrop}{$-4.91$} \\
& & Ours & \textbf{47.00} & \textbf{62.93} & \textbf{80.81} & \textbf{77.12} & 76.33 & \textbf{68.84} & \textcolor{fpdrop}{$\mathbf{-4.36}$} \\
\bottomrule
\end{tabular}
\end{table}

\begin{table}[!t]
\centering
\caption{InternVL2-26B under W4A8. NoCaps-lite: CIDEr $\times 100$.}
\label{tab:internvl26b}
\small
\setlength{\tabcolsep}{3pt}
\begin{adjustbox}{max width=\linewidth}
\begin{tabular}{lrrrrrr}
\toprule
Method & MMMU & ScienceQA & AI2D-lite & NoCaps-lite & Avg. & $\Delta$FP \\
\midrule
FP & 47.11 & 97.27 & 83.00 & 87.74 & 78.78 & --- \\
\specialrule{\lightrulewidth}{0pt}{0pt}
MBQ & 44.67 & 96.93 & 78.60 & 77.48 & 74.42 & \textcolor{fpdrop}{$-4.36$} \\
QIG & 44.00 & 96.63 & 78.40 & \textbf{79.66} & 74.67 & \textcolor{fpdrop}{$-4.11$} \\
Ours & \textbf{44.89} & \textbf{96.98} & \textbf{79.00} & 78.98 & \textbf{74.96} & \textcolor{fpdrop}{$\mathbf{-3.82}$} \\
\bottomrule
\end{tabular}
\end{adjustbox}
\end{table}

\section{Experiments}
\subsection{Experimental Setup}

\paragraph{Implementation Details.}
Following prior work~\citep{mbq,qig}, we use group-wise affine weight quantization and symmetric per-token activation quantization, and evaluate all methods under W3A16 and W4A8 settings.
Experiments use a single NVIDIA RTX A6000 GPU (48GB), except for InternVL2-8B~\citep{internvl} evaluation on MMMU~\citep{mmmu} and InternVL2-26B evaluation, which use NVIDIA DGX Spark for its larger unified memory.

\paragraph{Datasets and Models.}
We use the improved COCO Caption dataset~\citep{chen2015coco} from ShareGPT4V~\citep{chen2024sharegpt4v} for calibration, sampling 64 image--caption pairs formatted with each LVLM's prompt template; as this calibration set is already small, we use all samples for both quantization effect estimation and CWE optimization.
Following LMMs-Eval~\citep{zhang2025lmms}, we evaluate on MMMU~\citep{mmmu} and ScienceQA~\citep{scienceqa} for visual reasoning, VizWiz~\citep{vizwiz} for real-world perception, and ChartQA~\citep{chartqa} and AI2D~\citep{ai2d} for visual understanding.
We benchmark LLaVA-OneVision-7B~\citep{llavaov}, InternVL2-8B~\citep{internvl}, and Qwen2-VL-7B~\citep{qwen2}. For InternVL2-26B, we additionally evaluate image captioning on NoCaps-lite~\citep{agrawal2019nocaps}.
\paragraph{Baselines.}
We compare with round-to-nearest (RTN), Q-VLM~\citep{qvlm}, Modality-Balanced Quantization (MBQ)~\citep{mbq}, and Quantization-Aware Integrated Gradients (QIG)~\citep{qig}. For Q-VLM, MBQ, and QIG, we follow the official open-source implementations. Since Q-VLM is designed for LLaVA-style architectures, we adapt and re-implement it to ensure compatibility with all evaluated models.

\subsection{Main Results}
Table~\ref{tab:main_results} and Figure~\ref{fig:performance_comparison} show that Balanced Fitting outperforms existing PTQ methods in five-task average across all models and both precisions, indicating consistent performance preservation across diverse tasks. On MMMU~\citep{mmmu}, which requires multidisciplinary knowledge and visual reasoning, it reduces the accuracy drop by $2.43\times$ on average relative to the strongest competing baseline. Table~\ref{tab:internvl26b} extends this comparison to the larger InternVL2-26B under W4A8, where our method also leads in average performance.

\begin{figure}[!t]
    \centering
    \includegraphics[width=\linewidth]{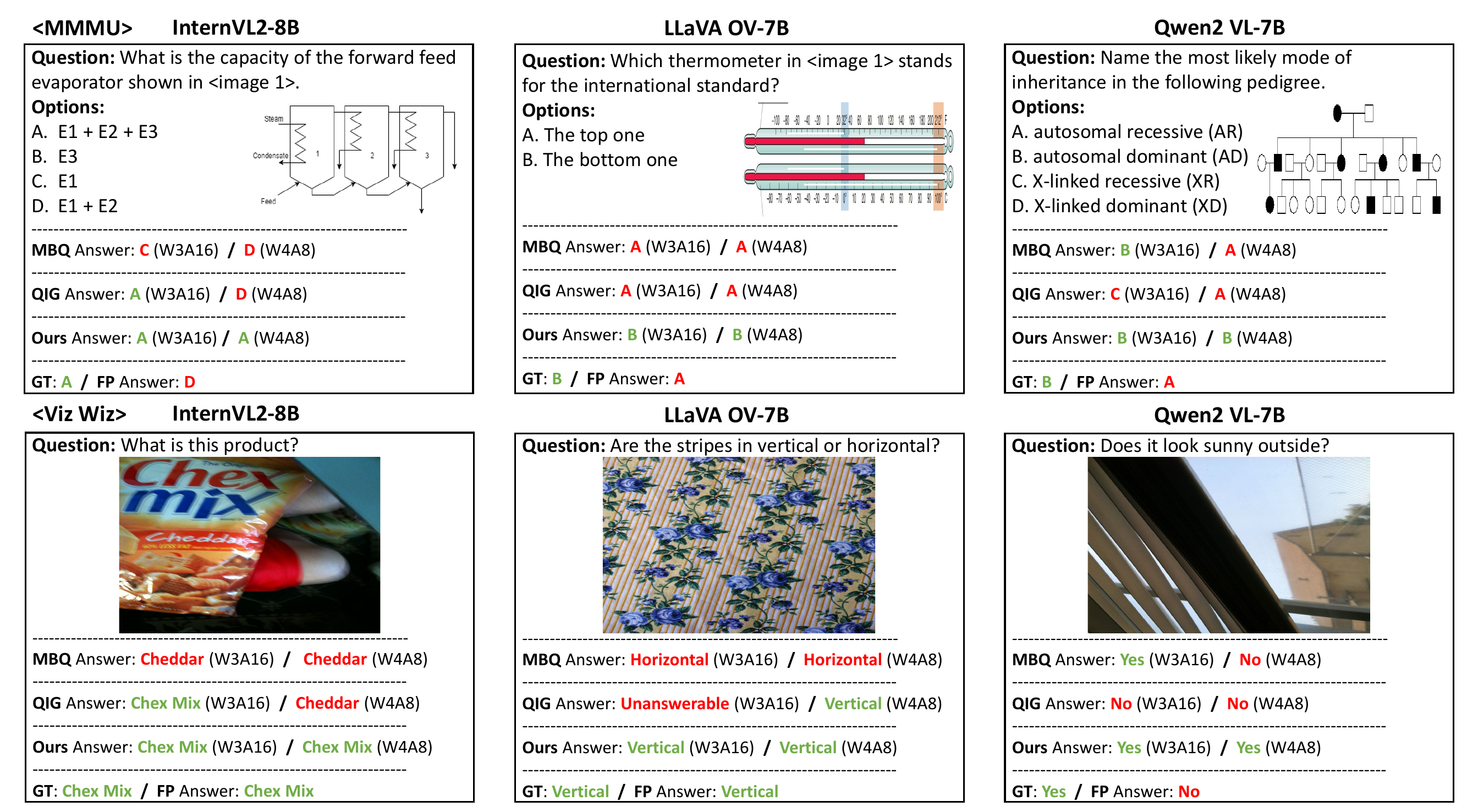}
    \caption{Qualitative comparison on MMMU and VizWiz.}
    \label{fig:qualitative_results}
\end{figure}

Figure~\ref{fig:qualitative_results} presents qualitative comparisons on MMMU and VizWiz. Balanced Fitting preserves correct FP answers while correcting erroneous predictions reproduced by MBQ and QIG, illustrating a balance between retaining useful FP behavior and allowing beneficial deviations.

\begin{table}[!htbp]
\centering
\caption{Ablation on the gain-discount coefficient ($\alpha$) and conflict coefficient ($\beta$) under the W4A8 setting.
Each cell reports MMMU / 5-task Avg.}
\label{tab:alpha_beta_ablation}

\begin{adjustbox}{width=\linewidth}
\begin{tabular}{lccc|ccc}
\toprule
Model
& \multicolumn{3}{c|}{Gain-discount coefficient ($\alpha$)}
& \multicolumn{3}{c}{Conflict coefficient ($\beta$)} \\
\cmidrule(lr){2-4}\cmidrule(lr){5-7}
& $0$ & $0.5$ & $1.0$
& $0$ & $0.5$ & $1.0$ \\
\midrule
Qwen2-VL-7B
& 46.67 / 68.70 & 47.00 / \textbf{68.84} & \textbf{47.33} / 68.83
& 45.67 / \textbf{68.98} & \textbf{47.00} / 68.84 & 46.22 / 68.59 \\
LLaVA-OV-7B
& 43.67 / 69.06 & \textbf{45.22 / 69.72} & 41.78 / 68.69
& 43.67 / 69.37 & \textbf{45.22 / 69.72} & 44.56 / 69.18 \\
InternVL2-8B
& 43.22 / 71.16 & \textbf{46.33} / 71.84 & 45.56 / \textbf{71.87}
& 45.44 / 71.44 & \textbf{46.33 / 71.84} & 44.67 / 71.44 \\
\bottomrule
\end{tabular}
\end{adjustbox}
\end{table}

\begin{table}[!htbp]
\centering
\caption{Uniform vs. adaptive grid allocation. Adap. ($k$) indicates mean grid size $k$.}
\label{tab:uniform_grid_all_models}

\begin{adjustbox}{width=\textwidth}
\begin{tabular}{llcccccccccccc}
\toprule
& & \multicolumn{4}{c}{InternVL2-8B} & \multicolumn{4}{c}{LLaVA-OV-7B} & \multicolumn{4}{c}{Qwen2-VL-7B} \\
\cmidrule(lr){3-6}\cmidrule(lr){7-10}\cmidrule(l){11-14}
Precision & Method
& Grid & Loss $\downarrow$ & MMMU $\uparrow$ & Avg. $\uparrow$
& Grid & Loss $\downarrow$ & MMMU $\uparrow$ & Avg. $\uparrow$
& Grid & Loss $\downarrow$ & MMMU $\uparrow$ & Avg. $\uparrow$ \\
\midrule
W3A16
& Low
& 18 & 2.34 & 46.4 & 72.4
& 16 & \textbf{0.82} & 42.6 & 70.9
& 20 & 0.85 & 47.2 & 70.5 \\
& Mean
& 23 & \textbf{1.85} & 46.6 & 72.3
& 27 & 0.96 & 44.8 & 70.6
& 22 & 0.85 & 47.4 & 70.4 \\
& High
& 28 & 1.94 & 45.9 & 72.1
& 32 & 0.94 & 45.7 & 70.8
& 24 & \textbf{0.79} & 46.6 & 70.0 \\
& Ours
& Adap. (23) & 2.34 & \textbf{47.4} & \textbf{72.6}
& Adap. (27) & 0.90 & \textbf{46.1} & \textbf{71.5}
& Adap. (22) & 0.87 & \textbf{48.0} & \textbf{70.6} \\
\midrule
W4A8
& Low
& 17 & \textbf{0.28} & 44.4 & 71.3
& 15 & 0.32 & 43.4 & 69.1
& 16 & 0.27 & 46.4 & \textbf{68.8} \\
& Mean
& 21 & 0.41 & 44.6 & 71.2
& 18 & 0.38 & 45.0 & 69.3
& 19 & \textbf{0.25} & 45.3 & 67.8 \\
& High
& 25 & 0.37 & 45.0 & 71.3
& 23 & \textbf{0.30} & 43.1 & 68.9
& 23 & \textbf{0.25} & 46.4 & 68.7 \\
& Ours
& Adap. (21) & 0.32 & \textbf{46.3} & \textbf{71.8}
& Adap. (18) & 0.33 & \textbf{45.2} & \textbf{69.7}
& Adap. (19) & 0.28 & \textbf{47.0} & \textbf{68.8} \\
\bottomrule
\end{tabular}
\end{adjustbox}
\end{table}

\subsection{Ablation Study and Analysis}
To examine the effect of each allocation component, we vary the gain-discount coefficient $\alpha$ in \eqref{eq:importance_score} and the conflict coefficient $\beta$ in \eqref{eq:conflict_score}. These coefficients control how strongly beneficial quantization effects are discounted and conflicting component responses are emphasized, respectively. Table~\ref{tab:alpha_beta_ablation} reports the resulting performance. $\alpha=\beta=0.5$ yields competitive performance across models and metrics.

We next examine whether adaptive allocation improves on uniform budgets, comparing reconstruction loss and downstream performance in Table~\ref{tab:uniform_grid_all_models}. Adaptive allocation consistently achieves superior performance across all models and settings. 
Notably, uniformly allocating high budgets can underperform configurations with lower budgets, indicating that excessive fitting can be detrimental. 
Moreover, we observe that models with lower reconstruction loss do not necessarily achieve better downstream performance, highlighting the limitation of reconstruction-based optimization.

\begin{figure}[!t]
    \centering
    \includegraphics[width=\linewidth]{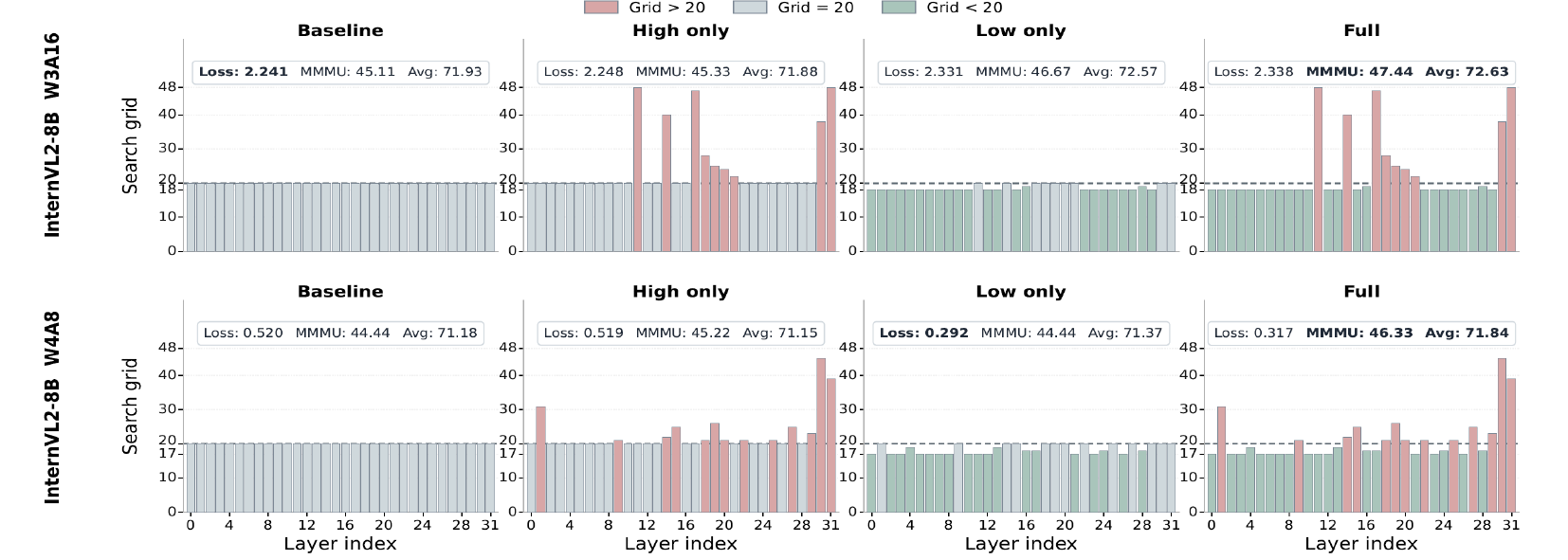}
    \caption{Layer-wise budget allocation and downstream performance on InternVL2-8B.}
    \label{fig:abl_decomp}
\end{figure}

\begin{table}[!t]
\centering
\caption{Quantization effect (QE) stability across disjoint calibration subsets and an independent held-out subset under W3A16 and W4A8. Parentheses give agreement counts.}
\label{tab:qe_subset_stability}
\small
\setlength{\tabcolsep}{3pt}
\begin{adjustbox}{max width=\linewidth}
\begin{tabular}{lccccc}
\toprule
QE-ranked subset & Count & QE mass & 3-subset sign agreement & Held-out agreement & Beneficial held-out agreement \\
\midrule
Top 50\% & 176 & 89.90\% & 87.5\% (154/176) & 93.2\% (164/176) & 89.8\% (44/49) \\
Top 25\% & 88 & 73.90\% & 94.3\% (83/88) & 100\% (88/88) & 100\% (23/23) \\
Top 10\% & 38 & 57.00\% & 100\% (38/38) & 100\% (38/38) & 100\% (10/10) \\
\bottomrule
\end{tabular}
\end{adjustbox}
\end{table}

\begin{table}[!t]
\centering
\caption{Calibration-domain shift from COCO Caption to Flickr30k: W4A8 five-task average.}
\label{tab:flickr_calibration}
\small
\setlength{\tabcolsep}{3pt}
\begin{adjustbox}{max width=\linewidth}
\begin{tabular}{lrrrr}
\toprule
Model & FP & MBQ ($\Delta$FP) & QIG ($\Delta$FP) & Ours ($\Delta$FP) \\
\midrule
InternVL2 & 74.14 & 71.60 (\textcolor{fpdrop}{$-2.54$}) & 70.87 (\textcolor{fpdrop}{$-3.27$}) & \textbf{71.78} (\textcolor{fpdrop}{$\mathbf{-2.36}$}) \\
LLaVA-OV & 72.48 & 68.06 (\textcolor{fpdrop}{$-4.42$}) & 68.92 (\textcolor{fpdrop}{$-3.56$}) & \textbf{69.43} (\textcolor{fpdrop}{$\mathbf{-3.05}$}) \\
Qwen2-VL & 73.20 & 67.57 (\textcolor{fpdrop}{$-5.63$}) & 67.27 (\textcolor{fpdrop}{$-5.93$}) & \textbf{67.68} (\textcolor{fpdrop}{$\mathbf{-5.52}$}) \\
\bottomrule
\end{tabular}
\end{adjustbox}
\end{table}

Figure~\ref{fig:abl_decomp} compares refinement (High-only), coarsening (Low-only), and Balanced Fitting (Full), which combines both. Reconstruction loss fails to track downstream performance across these variants, making it an unreliable performance proxy. Full achieves the best average; Low-only also outperforms the uniform baseline, consistent with a regularization effect from constrained fitting.

To verify that quantization effects remain stable across samples, we compare their signs across three disjoint COCO-64 subsets and an independent held-out-64 subset, ranking components using calibration data only. Table~\ref{tab:qe_subset_stability} shows strong held-out sign agreement for high-magnitude effects, including beneficial effects. This suggests that the signals guiding refinement and coarsening are not specific to a single calibration subset, supporting their use for budget allocation. Appendix Figure~\ref{fig:sample_stability} further shows consistent layer-wise estimates across sample sizes.

To test robustness to a calibration-domain shift from COCO Caption to Flickr30k, we re-estimate quantization effects on 64 Flickr30k training pairs and apply the unchanged W4A8 allocation rule without retuning. Table~\ref{tab:flickr_calibration} shows that Ours retains the highest five-task average across all three models, supporting robustness to a change in calibration data. Additional controls, analyses, and costs are provided in the appendix.

\section{Conclusion}

In this work, we revisit post-training quantization (PTQ) for large vision-language models (LVLMs) and show that minimizing reconstruction loss on small calibration sets is insufficient for generalization. 
We propose Balanced Fitting, a quantization effect-based framework that allocates layer-wise budgets to control fitting strength, balancing precision and regularization across components. 
Extensive experiments demonstrate consistent improvements over prior PTQ methods and reveal the limitation of reconstruction loss as a proxy for downstream performance.

\bibliographystyle{iclr2027_conference}
\bibliography{references}

@article{qig,
  title={Fine-Grained Post-Training Quantization for Large Vision Language Models with Quantization-Aware Integrated Gradients},
  author={Xiang, Ziwei and Zeng, Fanhu and Fang, Hongjian and Wang, Rui-Qi and Chen, Renxing and Zhu, Yanan and Chen, Yi and Yang, Peipei and Zhang, Xu-Yao},
  journal={arXiv preprint arXiv:2603.17809},
  year={2026}
}

@inproceedings{mbq,
  title={Mbq: Modality-balanced quantization for large vision-language models},
  author={Li, Shiyao and Hu, Yingchun and Ning, Xuefei and Liu, Xihui and Hong, Ke and Jia, Xiaotao and Li, Xiuhong and Yan, Yaqi and Ran, Pei and Dai, Guohao and others},
  booktitle={Proceedings of the Computer Vision and Pattern Recognition Conference},
  pages={4167--4177},
  year={2025}
}

@article{qvlm,
  title={Q-vlm: Post-training quantization for large vision-language models},
  author={Wang, Changyuan and Wang, Ziwei and Xu, Xiuwei and Tang, Yansong and Zhou, Jie and Lu, Jiwen},
  journal={Advances in Neural Information Processing Systems},
  volume={37},
  pages={114553--114573},
  year={2024}
}

@article{liu2023visual,
  title={Visual instruction tuning},
  author={Liu, Haotian and Li, Chunyuan and Wu, Qingyang and Lee, Yong Jae},
  journal={Advances in neural information processing systems},
  volume={36},
  pages={34892--34916},
  year={2023}
}

@article{qwen2,
  title={Qwen2-vl: Enhancing vision-language model's perception of the world at any resolution},
  author={Wang, Peng and Bai, Shuai and Tan, Sinan and Wang, Shijie and Fan, Zhihao and Bai, Jinze and Chen, Keqin and Liu, Xuejing and Wang, Jialin and Ge, Wenbin and others},
  journal={arXiv preprint arXiv:2409.12191},
  year={2024}
}

@article{llavaov,
  title={Llava-onevision: Easy visual task transfer},
  author={Li, Bo and Zhang, Yuanhan and Guo, Dong and Zhang, Renrui and Li, Feng and Zhang, Hao and Zhang, Kaichen and Zhang, Peiyuan and Li, Yanwei and Liu, Ziwei and others},
  journal={arXiv preprint arXiv:2408.03326},
  year={2024}
}

@inproceedings{internvl,
  title={Internvl: Scaling up vision foundation models and aligning for generic visual-linguistic tasks},
  author={Chen, Zhe and Wu, Jiannan and Wang, Wenhai and Su, Weijie and Chen, Guo and Xing, Sen and Zhong, Muyan and Zhang, Qinglong and Zhu, Xizhou and Lu, Lewei and others},
  booktitle={Proceedings of the IEEE/CVF conference on computer vision and pattern recognition},
  pages={24185--24198},
  year={2024}
}

@article{alayrac2022flamingo,
  title={Flamingo: a visual language model for few-shot learning},
  author={Alayrac, Jean-Baptiste and Donahue, Jeff and Luc, Pauline and Miech, Antoine and Barr, Iain and Hasson, Yana and Lenc, Karel and Mensch, Arthur and Millican, Katherine and Reynolds, Malcolm and others},
  journal={Advances in neural information processing systems},
  volume={35},
  pages={23716--23736},
  year={2022}
}

@inproceedings{li2023blip,
  title={Blip-2: Bootstrapping language-image pre-training with frozen image encoders and large language models},
  author={Li, Junnan and Li, Dongxu and Savarese, Silvio and Hoi, Steven},
  booktitle={International conference on machine learning},
  pages={19730--19742},
  year={2023},
  organization={PMLR}
}

@article{rao2021dynamicvit,
  title={Dynamicvit: Efficient vision transformers with dynamic token sparsification},
  author={Rao, Yongming and Zhao, Wenliang and Liu, Benlin and Lu, Jiwen and Zhou, Jie and Hsieh, Cho-Jui},
  journal={Advances in neural information processing systems},
  volume={34},
  pages={13937--13949},
  year={2021}
}

@article{bolya2022token,
  title={Token merging: Your vit but faster},
  author={Bolya, Daniel and Fu, Cheng-Yang and Dai, Xiaoliang and Zhang, Peizhao and Feichtenhofer, Christoph and Hoffman, Judy},
  journal={arXiv preprint arXiv:2210.09461},
  year={2022}
}

@article{hinton2015distilling,
  title={Distilling the knowledge in a neural network},
  author={Hinton, Geoffrey and Vinyals, Oriol and Dean, Jeff},
  journal={arXiv preprint arXiv:1503.02531},
  year={2015}
}

@inproceedings{li2023distilling,
  title={Distilling large vision-language model with out-of-distribution generalizability},
  author={Li, Xuanlin and Fang, Yunhao and Liu, Minghua and Ling, Zhan and Tu, Zhuowen and Su, Hao},
  booktitle={Proceedings of the IEEE/CVF International Conference on Computer Vision},
  pages={2492--2503},
  year={2023}
}

@article{gptq,
  title={Gptq: Accurate post-training quantization for generative pre-trained transformers},
  author={Frantar, Elias and Ashkboos, Saleh and Hoefler, Torsten and Alistarh, Dan},
  journal={arXiv preprint arXiv:2210.17323},
  year={2022}
}

@article{awq,
  title={Awq: Activation-aware weight quantization for on-device llm compression and acceleration},
  author={Lin, Ji and Tang, Jiaming and Tang, Haotian and Yang, Shang and Chen, Wei-Ming and Wang, Wei-Chen and Xiao, Guangxuan and Dang, Xingyu and Gan, Chuang and Han, Song},
  journal={Proceedings of machine learning and systems},
  volume={6},
  pages={87--100},
  year={2024}
}

@inproceedings{smoothquant,
  title={Smoothquant: Accurate and efficient post-training quantization for large language models},
  author={Xiao, Guangxuan and Lin, Ji and Seznec, Mickael and Wu, Hao and Demouth, Julien and Han, Song},
  booktitle={International conference on machine learning},
  pages={38087--38099},
  year={2023},
  organization={PMLR}
}

@inproceedings{nagel2020up,
  title={Up or down? adaptive rounding for post-training quantization},
  author={Nagel, Markus and Amjad, Rana Ali and Van Baalen, Mart and Louizos, Christos and Blankevoort, Tijmen},
  booktitle={International conference on machine learning},
  pages={7197--7206},
  year={2020},
  organization={PMLR}
}

@inproceedings{jacob2018quantization,
  title={Quantization and training of neural networks for efficient integer-arithmetic-only inference},
  author={Jacob, Benoit and Kligys, Skirmantas and Chen, Bo and Zhu, Menglong and Tang, Matthew and Howard, Andrew and Adam, Hartwig and Kalenichenko, Dmitry},
  booktitle={Proceedings of the IEEE conference on computer vision and pattern recognition},
  pages={2704--2713},
  year={2018}
}

@article{zhou2016dorefa,
  title={Dorefa-net: Training low bitwidth convolutional neural networks with low bitwidth gradients},
  author={Zhou, Shuchang and Wu, Yuxin and Ni, Zekun and Zhou, Xinyu and Wen, He and Zou, Yuheng},
  journal={arXiv preprint arXiv:1606.06160},
  year={2016}
}

@article{chen2021quantization,
  title={Quantization of deep neural networks for accurate edge computing},
  author={Chen, Wentao and Qiu, Hailong and Zhuang, Jian and Zhang, Chutong and Hu, Yu and Lu, Qing and Wang, Tianchen and Shi, Yiyu and Huang, Meiping and Xu, Xiaowe},
  journal={ACM Journal on Emerging Technologies in Computing Systems (JETC)},
  volume={17},
  number={4},
  pages={1--11},
  year={2021},
  publisher={ACM New York, NY}
}

@article{liang2021pruning,
  title={Pruning and quantization for deep neural network acceleration: A survey},
  author={Liang, Tailin and Glossner, John and Wang, Lei and Shi, Shaobo and Zhang, Xiaotong},
  journal={Neurocomputing},
  volume={461},
  pages={370--403},
  year={2021},
  publisher={Elsevier}
}

@article{kuzmin2023pruning,
  title={Pruning vs quantization: Which is better?},
  author={Kuzmin, Andrey and Nagel, Markus and Van Baalen, Mart and Behboodi, Arash and Blankevoort, Tijmen},
  journal={Advances in neural information processing systems},
  volume={36},
  pages={62414--62427},
  year={2023}
}

@inproceedings{wang2020differentiable,
  title={Differentiable joint pruning and quantization for hardware efficiency},
  author={Wang, Ying and Lu, Yadong and Blankevoort, Tijmen},
  booktitle={European Conference on Computer Vision},
  pages={259--277},
  year={2020},
  organization={Springer}
}

@inproceedings{mmmu,
  title={Mmmu: A massive multi-discipline multimodal understanding and reasoning benchmark for expert agi},
  author={Yue, Xiang and Ni, Yuansheng and Zhang, Kai and Zheng, Tianyu and Liu, Ruoqi and Zhang, Ge and Stevens, Samuel and Jiang, Dongfu and Ren, Weiming and Sun, Yuxuan and others},
  booktitle={Proceedings of the IEEE/CVF conference on computer vision and pattern recognition},
  pages={9556--9567},
  year={2024}
}

@inproceedings{vizwiz,
  title={Vizwiz grand challenge: Answering visual questions from blind people},
  author={Gurari, Danna and Li, Qing and Stangl, Abigale J and Guo, Anhong and Lin, Chi and Grauman, Kristen and Luo, Jiebo and Bigham, Jeffrey P},
  booktitle={Proceedings of the IEEE conference on computer vision and pattern recognition},
  pages={3608--3617},
  year={2018}
}

@article{scienceqa,
  title={Learn to explain: Multimodal reasoning via thought chains for science question answering},
  author={Lu, Pan and Mishra, Swaroop and Xia, Tanglin and Qiu, Liang and Chang, Kai-Wei and Zhu, Song-Chun and Tafjord, Oyvind and Clark, Peter and Kalyan, Ashwin},
  journal={Advances in neural information processing systems},
  volume={35},
  pages={2507--2521},
  year={2022}
}

@inproceedings{clip,
  title={Learning transferable visual models from natural language supervision},
  author={Radford, Alec and Kim, Jong Wook and Hallacy, Chris and Ramesh, Aditya and Goh, Gabriel and Agarwal, Sandhini and Sastry, Girish and Askell, Amanda and Mishkin, Pamela and Clark, Jack and others},
  booktitle={International conference on machine learning},
  pages={8748--8763},
  year={2021},
  organization={PmLR}
}

@inproceedings{chartqa,
  title={ChartQA: A Benchmark for Question Answering about Charts with Visual and Logical Reasoning},
  author={Masry, Ahmed and Long, Do and Tan, Jiaqi and Joty, Shafiq and Hoque, Enamul},
  booktitle={Findings of the Association for Computational Linguistics: ACL 2022},
  pages={2263--2279},
  year={2022}
}

@inproceedings{ai2d,
  title={Are you smarter than a sixth grader? textbook question answering for multimodal machine comprehension},
  author={Kembhavi, Aniruddha and Seo, Minjoon and Schwenk, Dustin and Choi, Jonghyun and Farhadi, Ali and Hajishirzi, Hannaneh},
  booktitle={Proceedings of the IEEE Conference on Computer Vision and Pattern recognition},
  pages={4999--5007},
  year={2017}
}

@article{brown2020language,
  title={Language models are few-shot learners},
  author={Brown, Tom and Mann, Benjamin and Ryder, Nick and Subbiah, Melanie and Kaplan, Jared D and Dhariwal, Prafulla and Neelakantan, Arvind and Shyam, Pranav and Sastry, Girish and Askell, Amanda and others},
  journal={Advances in neural information processing systems},
  volume={33},
  pages={1877--1901},
  year={2020}
}

@article{chowdhery2023palm,
  title={Palm: Scaling language modeling with pathways},
  author={Chowdhery, Aakanksha and Narang, Sharan and Devlin, Jacob and Bosma, Maarten and Mishra, Gaurav and Roberts, Adam and Barham, Paul and Chung, Hyung Won and Sutton, Charles and Gehrmann, Sebastian and others},
  journal={Journal of machine learning research},
  volume={24},
  number={240},
  pages={1--113},
  year={2023}
}

@article{touvron2023llama,
  title={Llama: Open and efficient foundation language models},
  author={Touvron, Hugo and Lavril, Thibaut and Izacard, Gautier and Martinet, Xavier and Lachaux, Marie-Anne and Lacroix, Timoth{\'e}e and Rozi{\`e}re, Baptiste and Goyal, Naman and Hambro, Eric and Azhar, Faisal and others},
  journal={arXiv preprint arXiv:2302.13971},
  year={2023}
}

@article{dosovitskiy2020image,
  title={An image is worth 16x16 words: Transformers for image recognition at scale},
  author={Dosovitskiy, Alexey and Beyer, Lucas and Kolesnikov, Alexander and Weissenborn, Dirk and Zhai, Xiaohua and Unterthiner, Thomas and Dehghani, Mostafa and Minderer, Matthias and Heigold, Georg and Gelly, Sylvain and others},
  journal={arXiv preprint arXiv:2010.11929},
  year={2020}
}

@article{liu2021post,
  title={Post-training quantization for vision transformer},
  author={Liu, Zhenhua and Wang, Yunhe and Han, Kai and Zhang, Wei and Ma, Siwei and Gao, Wen},
  journal={Advances in Neural Information Processing Systems},
  volume={34},
  pages={28092--28103},
  year={2021}
}

@inproceedings{yuan2022ptq4vit,
  title={Ptq4vit: Post-training quantization for vision transformers with twin uniform quantization},
  author={Yuan, Zhihang and Xue, Chenhao and Chen, Yiqi and Wu, Qiang and Sun, Guangyu},
  booktitle={European conference on computer vision},
  pages={191--207},
  year={2022},
  organization={Springer}
}

@inproceedings{li2023vit,
  title={I-vit: Integer-only quantization for efficient vision transformer inference},
  author={Li, Zhikai and Gu, Qingyi},
  booktitle={Proceedings of the IEEE/CVF International Conference on Computer Vision},
  pages={17065--17075},
  year={2023}
}

@article{lin2021fq,
  title={Fq-vit: Post-training quantization for fully quantized vision transformer},
  author={Lin, Yang and Zhang, Tianyu and Sun, Peiqin and Li, Zheng and Zhou, Shuchang},
  journal={arXiv preprint arXiv:2111.13824},
  year={2021}
}

@article{yao2022zeroquant,
  title={Zeroquant: Efficient and affordable post-training quantization for large-scale transformers},
  author={Yao, Zhewei and Yazdani Aminabadi, Reza and Zhang, Minjia and Wu, Xiaoxia and Li, Conglong and He, Yuxiong},
  journal={Advances in neural information processing systems},
  volume={35},
  pages={27168--27183},
  year={2022}
}

@article{wei2022outlier,
  title={Outlier suppression: Pushing the limit of low-bit transformer language models},
  author={Wei, Xiuying and Zhang, Yunchen and Zhang, Xiangguo and Gong, Ruihao and Zhang, Shanghang and Zhang, Qi and Yu, Fengwei and Liu, Xianglong},
  journal={Advances in Neural Information Processing Systems},
  volume={35},
  pages={17402--17414},
  year={2022}
}

@inproceedings{liu2023noisyquant,
  title={Noisyquant: Noisy bias-enhanced post-training activation quantization for vision transformers},
  author={Liu, Yijiang and Yang, Huanrui and Dong, Zhen and Keutzer, Kurt and Du, Li and Zhang, Shanghang},
  booktitle={Proceedings of the IEEE/CVF conference on computer vision and pattern recognition},
  pages={20321--20330},
  year={2023}
}

@article{banner2019post,
  title={Post training 4-bit quantization of convolutional networks for rapid-deployment},
  author={Banner, Ron and Nahshan, Yury and Soudry, Daniel},
  journal={Advances in neural information processing systems},
  volume={32},
  year={2019}
}

@inproceedings{chen2024sharegpt4v,
  title={Sharegpt4v: Improving large multi-modal models with better captions},
  author={Chen, Lin and Li, Jinsong and Dong, Xiaoyi and Zhang, Pan and He, Conghui and Wang, Jiaqi and Zhao, Feng and Lin, Dahua},
  booktitle={European Conference on Computer Vision},
  pages={370--387},
  year={2024},
  organization={Springer}
}

@article{chen2015coco,
  title={Microsoft COCO Captions: Data Collection and Evaluation Server},
  author={Chen, Xinlei and Fang, Hao and Lin, Tsung-Yi and Vedantam, Ramakrishna and Gupta, Saurabh and Doll{\'a}r, Piotr and Zitnick, C. Lawrence},
  journal={arXiv preprint arXiv:1504.00325},
  year={2015}
}

@inproceedings{zhang2025lmms,
  title={Lmms-eval: Reality check on the evaluation of large multimodal models},
  author={Zhang, Kaichen and Li, Bo and Zhang, Peiyuan and Pu, Fanyi and Cahyono, Joshua Adrian and Hu, Kairui and Liu, Shuai and Zhang, Yuanhan and Yang, Jingkang and Li, Chunyuan and others},
  booktitle={Findings of the Association for Computational Linguistics: NAACL 2025},
  pages={881--916},
  year={2025}
}

@inproceedings{agrawal2019nocaps,
  title={Nocaps: Novel object captioning at scale},
  author={Agrawal, Harsh and Desai, Karan and Wang, Yufei and Chen, Xinlei and Jain, Rishabh and Johnson, Mark and Batra, Dhruv and Parikh, Devi and Lee, Stefan and Anderson, Peter},
  booktitle={2019 IEEE/CVF International Conference on Computer Vision (ICCV)},
  pages={8947--8956},
  year={2019},
  organization={IEEE}
}

\appendix

\section{Additional Analysis}

\begin{figure}
    \centering
    \includegraphics[width=\linewidth]{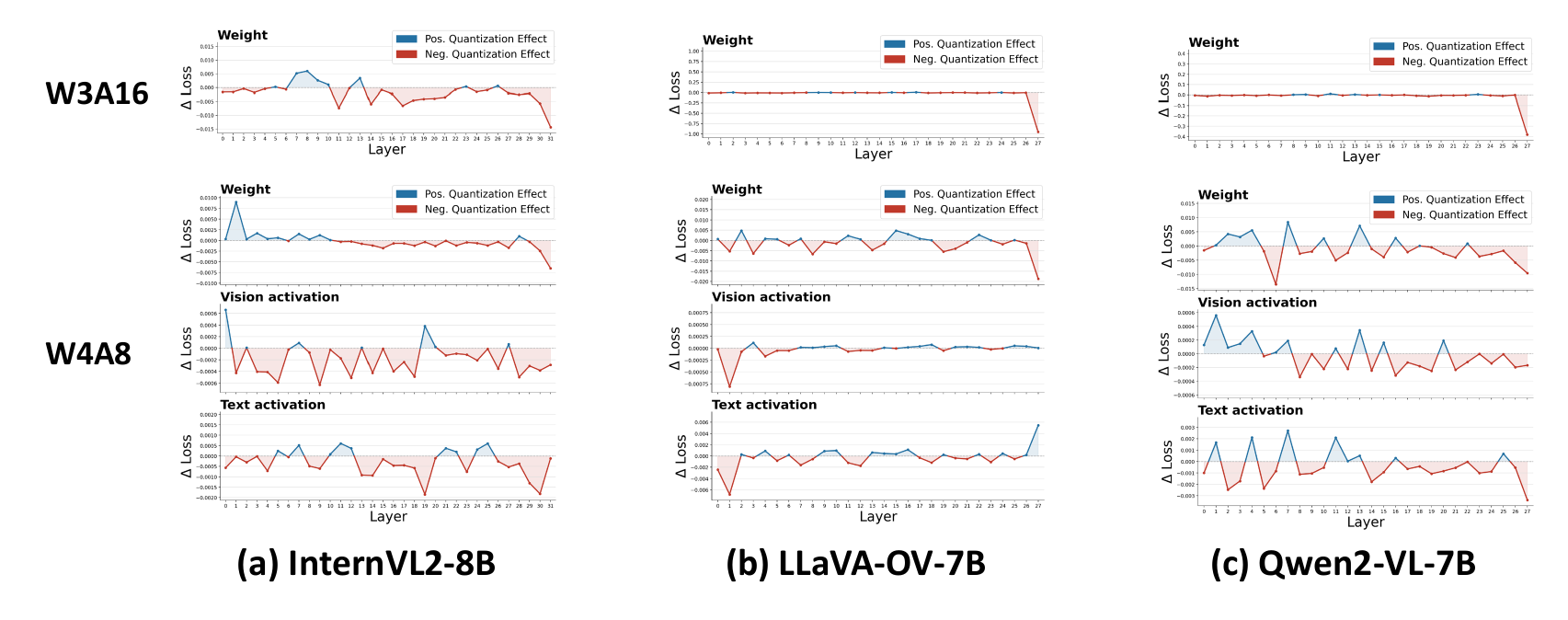}
    \caption{Component-wise quantization effects across layers for three vision-language models (InternVL2-8B, LLaVA-OV-7B, and Qwen2-VL-7B) under two quantization settings (W3A16 and W4A8). The results reveal heterogeneous sensitivity across weights and modalities, where some components benefit from quantization while others are more sensitive.}
    \label{fig:quantization_effects_appendix}
\end{figure}

\paragraph{Component-wise Quantization Effects Across Models and Settings.}
Figure~\ref{fig:quantization_effects_appendix} extends Figure~\ref{fig:motivation}(a) to InternVL2-8B~\citep{internvl}, LLaVA-OV-7B~\citep{llavaov}, and Qwen2-VL-7B~\citep{qwen2} under W3A16 and W4A8.

The sign and magnitude of quantization effects vary across layers, components, models, and bit-widths. Negative effects indicate increased calibration loss, whereas positive effects indicate reduced loss and suggest potential regularization benefits. The locations of sensitive and beneficial components differ across models and settings, supporting allocation based on the measured effects rather than a fixed layer-wise policy.

\begin{figure}
    \centering
    \includegraphics[width=\linewidth]{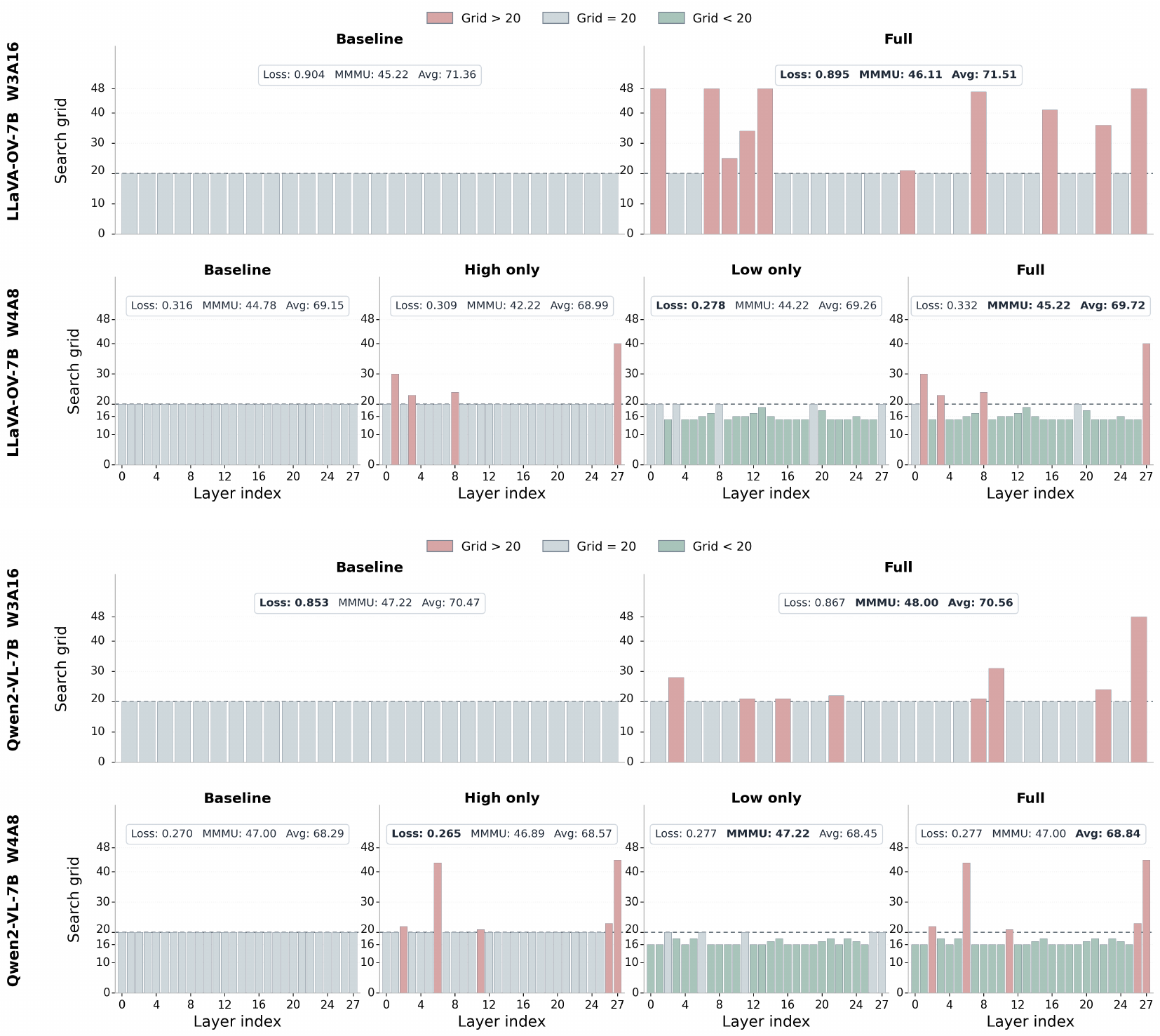}
    \caption{Layer-wise budget allocation and downstream performance on LLaVA-OV-7B and Qwen2-VL-7B. For the W3A16 setting, the minimum grid size is set to 20, making the high-only configuration equivalent to the full setting and the low-only configuration equivalent to the baseline; thus, they are excluded from comparison.}
    \label{fig:decomp_ablation_appendix}
\end{figure}

\paragraph{Effect of Balanced Budget Allocation on Downstream Performance.}
Figure~\ref{fig:decomp_ablation_appendix} extends the allocation ablation to LLaVA-OV-7B and Qwen2-VL-7B. Under W4A8, it compares uniform allocation, refinement alone (High-only), coarsening alone (Low-only), and Balanced Fitting (Full), allowing the contributions of refinement and coarsening to be examined separately. The results support combining both operations rather than relying on either alone. Under W3A16, the minimum grid size is 20, so High-only coincides with Full and Low-only coincides with the uniform baseline; only the distinct configurations are shown. This setting therefore evaluates refinement above the baseline budget rather than a separate coarsening effect.

\begin{figure}[!tbp]
    \centering
    \includegraphics[width=\linewidth]{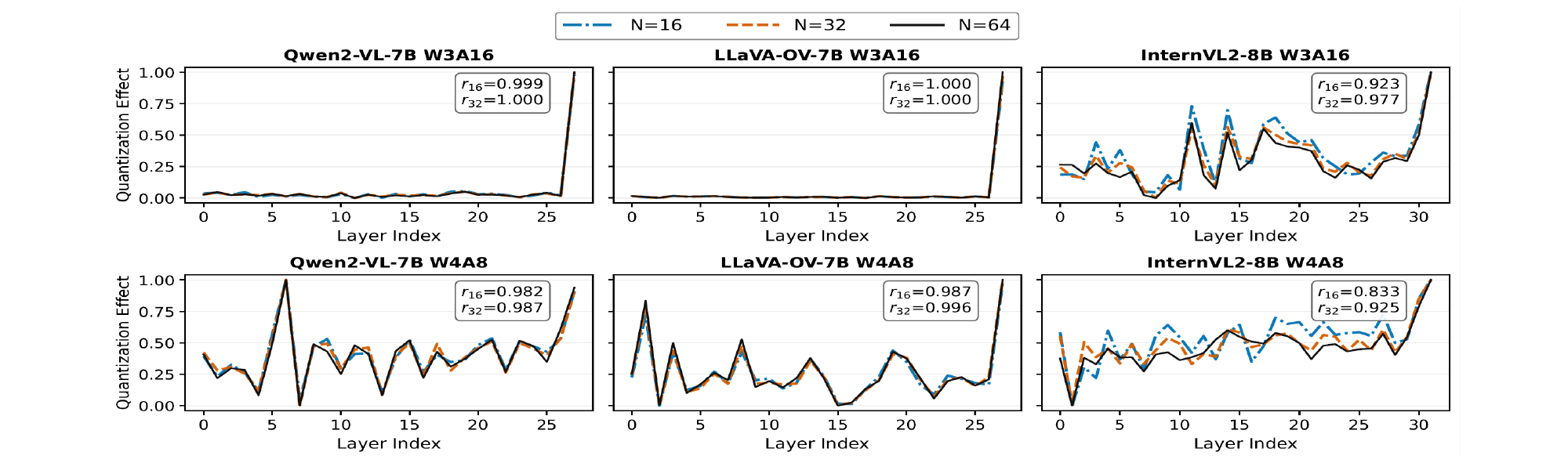}
    \caption{Stability of layer-wise quantization effects estimated from 16, 32, and 64 calibration samples on Qwen2-VL-7B, LLaVA-OV-7B, and InternVL2-8B under W3A16 (top) and W4A8 (bottom). W4A8 aggregates weight, visual-token, and textual-token effects. The reported Pearson correlations compare the 16- and 32-sample profiles with the 64-sample reference.}
    \label{fig:sample_stability}
\end{figure}

To examine how calibration-set size affects quantization effect estimation, Figure~\ref{fig:sample_stability} compares layer-wise profiles using 16, 32, and 64 samples across three LVLMs under W3A16 and W4A8. The profiles remain highly correlated across sample sizes, supporting reliable estimation with limited calibration data. This complements Table~\ref{tab:qe_subset_stability}, which tests sign agreement across disjoint calibration subsets and an independent held-out subset.

\section{Quantization Effect-Guided Allocation Rule}
\label{sec:auto_rule}

To reduce the manual effort in configuring Balanced Fitting, we provide a simple statistics-driven rule that maps quantization effect distributions to allocation settings. 
This rule selects allocation settings automatically and achieves performance close to empirically selected configurations.
The same formulas and normalization constants are used across models within each quantization mode.
The rule operates directly on summary statistics of quantization effects and is included as a practical supplementary tool.

\paragraph{Quantization effect statistics.}
For each layer $\ell$ and component $c$, let $s_c^{(\ell)}$ denote the quantization effect defined in \eqref{eq:quant_effect}. We decompose it into positive and negative parts:
\begin{equation}
s^{\mathrm{loss}}_{\ell,c} = \max(-s_c^{(\ell)}, 0),
\qquad
s^{\mathrm{gain}}_{\ell,c} = \max(s_c^{(\ell)}, 0).
\end{equation}

For W3A16, we use only the weight component:
\begin{equation}
H_\ell = s^{\mathrm{loss}}_{\ell,w},
\qquad
G_\ell = s^{\mathrm{gain}}_{\ell,w},
\qquad
H = \sum_{\ell} H_\ell,
\qquad
G = \sum_{\ell} G_\ell.
\end{equation}
We then define the gain ratio
\begin{equation}
r = \frac{G}{H+G},
\end{equation}
and the loss severity ratio
\begin{equation}
M = \frac{H}{\mathcal{L}_{\mathrm{task}}^{(\mathrm{FP})}},
\end{equation}
where $\mathcal{L}_{\mathrm{task}}^{(\mathrm{FP})}$ is the FP model's calibration-set task loss.

For W4A8, we aggregate all components:
\begin{equation}
H_\ell = \sum_{c \in \{w,v,t\}} s^{\mathrm{loss}}_{\ell,c},
\qquad
H = \sum_{\ell} H_\ell.
\end{equation}
We additionally define the activation-side loss share
\begin{equation}
A = \frac{\sum_{\ell} \left(s^{\mathrm{loss}}_{\ell,v}+s^{\mathrm{loss}}_{\ell,t}\right)}{H},
\end{equation}
and the support ratio
\begin{equation}
S = \frac{k_{0.9}}{L},
\end{equation}
where $k_{0.9}$ is the minimum number of layers needed to explain $90\%$ of the total loss mass and $L$ is the number of layers.

For both quantization modes, we measure concentration using the top $10\%$ of layers with positive loss mass.
Let $\mathcal{I}_{0.1}$ denote the set of layer indices corresponding to the largest $\lceil 0.1L_{+} \rceil$ values of $H_\ell$ among layers with $H_\ell>0$, where $L_{+}$ is the number of layers satisfying $H_\ell>0$.
We define
\begin{equation}
C_{0.1}
=
\frac{\sum_{\ell \in \mathcal{I}_{0.1}} H_\ell}{H}.
\end{equation}

\paragraph{Normalization.}
To place the raw statistics on a comparable scale, we normalize each statistic
$z$ into the unit interval:
\begin{equation}
\bar z =
\mathrm{clip}\!\left(
\frac{z-z_{\min}}{z_{\max}-z_{\min}},\,0,\,1
\right),
\end{equation}
where $z_{\min}$ and $z_{\max}$ are the normalization bounds. W3A16 uses $r\in[0.017,0.207]$, $C_{0.1}\in[0.375,0.915]$, and $M\in[0.065,1.371]$ (bounds shown rounded to three decimal places). For W4A8, we retain the zero origin of the ratios and round their upper bounds upward to one decimal: $A\in[0,0.8]$, $S\in[0,0.7]$, and $C_{0.1}\in[0,0.6]$. Values outside these ranges are clipped. These bounds and all coefficients below are shared across models within a quantization mode.

\paragraph{W3A16 rule.}
For W3A16, after normalizing $r$, $M$, and $C_{0.1}$ into $\bar r$, $\bar M$, and $\bar C$, we compute
\begin{align}
g_{\min}^{(\mathrm{w3a16})}
&=
\mathrm{round}\!\left(
g_{\mathrm{ref}}
-
\left(2+1.6(1-\bar r)\right)\bar r(1-\bar C)
\right), \\
g_{\mathrm{avg}}^{(\mathrm{w3a16})}
&=
g_{\mathrm{ref}} + 2
+ \mathrm{round}\!\left(5\,\bar C\,\bar M^3 + \bar r(1-\bar C)\right), \\
\gamma^{(\mathrm{w3a16})}
&=
\left[\,2.12+0.45\bar C\,\right]_{0.05},
\end{align}
where $[\cdot]_{0.05}$ denotes rounding to the nearest multiple of $0.05$.

\paragraph{W4A8 rule.}
For W4A8, after normalizing $A$, $S$, and $C_{0.1}$ into $\bar A$, $\bar S$, and $\bar C$, we compute
\begin{align}
g_{\min}^{(\mathrm{w4a8})}
&=
g_{\mathrm{ref}}-8+\mathrm{round}\!\left(2\bar A\bar S+9.5(1-\bar C)\right), \\
g_{\mathrm{avg}}^{(\mathrm{w4a8})}
&=
g_{\mathrm{ref}}-4+\mathrm{round}\!\left(9\bar A\bar S+\bar S(1-\bar C)\right), \\
\gamma^{(\mathrm{w4a8})}
&=
\left[\,1+0.9\bar C+0.6(1-\bar A)(1-\bar C)\,\right]_{0.05}.
\end{align}

Here, $g_{\mathrm{ref}}$ denotes the common reference grid budget. In our experiments, we follow prior work and set $g_{\mathrm{ref}}=20$. After rounding, we enforce $2\leq g_{\min}\leq g_{\mathrm{avg}}\leq48$; no-effect profiles use uniform 20-grid search.
Intuitively, for W4A8, $g_{\min}$ decreases when the loss profile is highly concentrated and increases when the loss mass is more diffuse or activation-heavy.
The target budget $g_{\mathrm{avg}}$ increases when activation-side loss is broader and stronger, while $\gamma$ is governed by concentration with an adjustment for diffuse low-activation cases.

\begin{figure}[t]
    \centering
    \includegraphics[width=\linewidth]{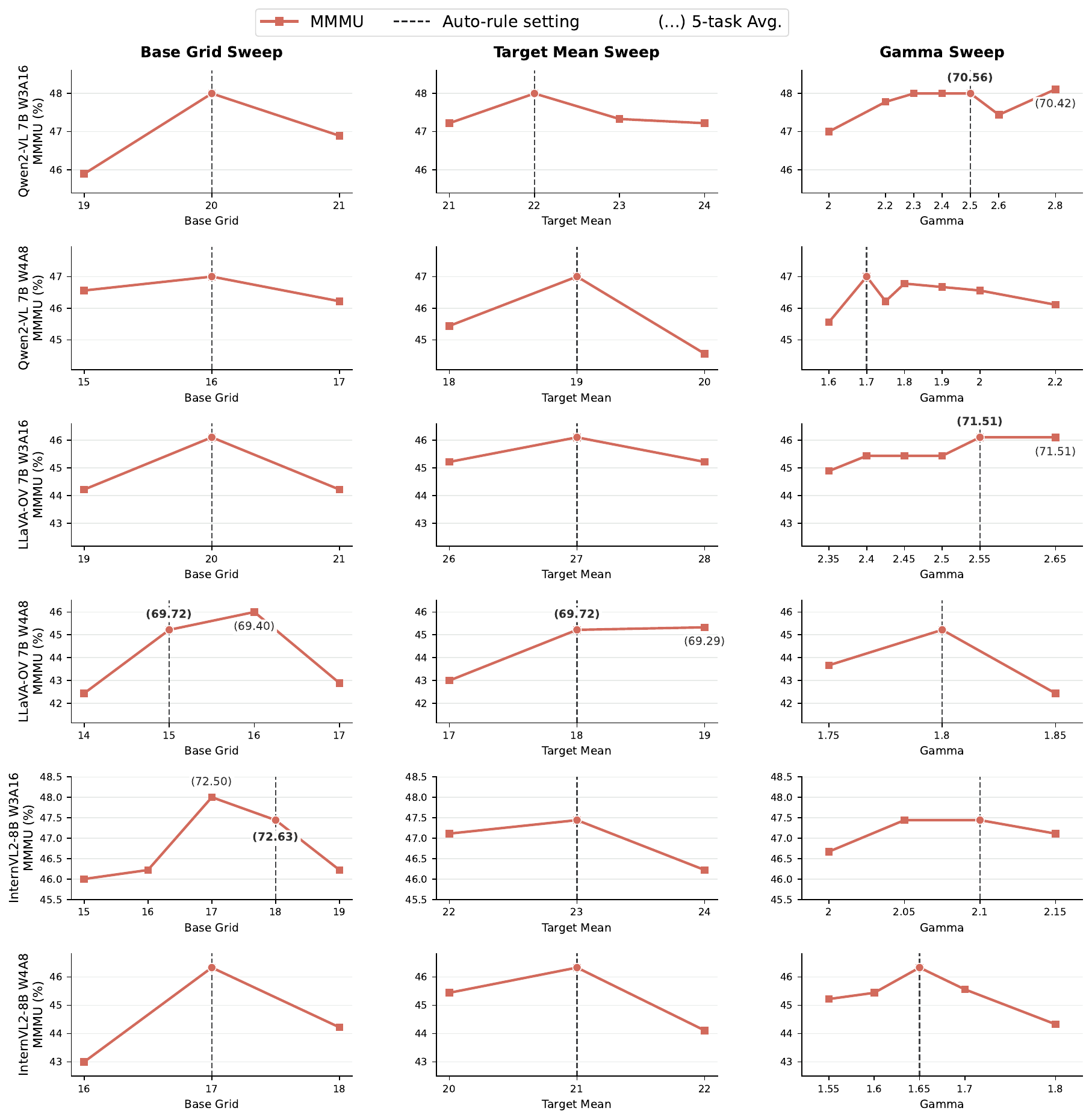}
    \caption{Analysis of the proposed Quantization Effect-Guided Allocation Rule. Each row corresponds to a model and setting, and each column sweeps one allocation dimension (base grid, target mean, or $\gamma$) while keeping the others fixed. Numbers in parentheses indicate the 5-task average performance, and curves report MMMU performance. The dashed vertical line indicates the reference allocation in the sweep.}
    \label{fig:autorule}
\end{figure}

Figure~\ref{fig:autorule} examines key allocation dimensions, including base grid, target mean, and $\gamma$. The selected settings are close to the best observed configurations, even when they do not always maximize MMMU.
\paragraph{Interpretation.}
The rule remains interpretable by linking allocation decisions to quantization effect statistics. 
For W3A16, the gain ratio and loss concentration jointly determine the minimum budget and allocation sharpness. 
For W4A8, the allocation is further influenced by the spread and magnitude of activation-side loss, leading to increased budgets when activation sensitivity is broader and stronger.

\section{Additional Comparative Evaluation}
\label{sec:additional_quantitative_evaluation}

\subsection{Evaluation settings}
InternVL2-26B uses COCO-64 on one NVIDIA DGX Spark (GB10, 128 GB unified memory). MMMU and ScienceQA use at most five image tiles instead of six for all methods, including FP. Its QE is measured directly, and the common rule reproduces $g_{\min}=14$, $g_{\mathrm{avg}}=21$, and $\gamma=1.8$. AI2D-lite and NoCaps-lite use the official fixed 500-example splits. For Flickr30k, we re-estimate quantization effects on 64 training pairs and apply the same allocation rule, normalization bounds, and coefficients without retuning, yielding $(g_{\min},g_{\mathrm{avg}},\gamma)=(16,25,1.65)$, $(16,18,1.90)$, and $(16,30,1.70)$ for InternVL2, LLaVA-OV, and Qwen2-VL, respectively. Calibration pairs do not overlap with the 1,000-image Karpathy test split.

\clearpage
\begin{table}[!t]
\centering
\caption{W3A16 per-task results with LLM-backbone AWQ and GPTQ; FP16 is the reference. Bold denotes the best quantized result.}
\label{tab:llm_backbone_baselines}
\small
\setlength{\tabcolsep}{3pt}
\begin{adjustbox}{max width=\linewidth}
\begin{tabular}{llrrrrrr}
\toprule
Model & Method & MMMU & VizWiz & ScienceQA & ChartQA & AI2D & Avg. \\
\midrule
\multirow{4}{*}{InternVL2} & FP16 & 48.11 & 60.67 & 97.07 & 82.48 & 82.35 & 74.14 \\
\cmidrule{2-8}
 & AWQ & 45.22 & 58.06 & 95.84 & 79.36 & \textbf{79.99} & 71.69 \\
 & GPTQ & 41.33 & \textbf{60.52} & 93.90 & 75.68 & 75.58 & 69.40 \\
 & Ours & \textbf{47.44} & 59.77 & \textbf{96.18} & \textbf{80.12} & 79.63 & \textbf{72.63} \\
\midrule
\multirow{4}{*}{LLaVA-OV} & FP16 & 46.56 & 58.71 & 95.84 & 80.00 & 81.28 & 72.48 \\
\cmidrule{2-8}
 & AWQ & 41.78 & 58.59 & \textbf{95.09} & 77.04 & 78.59 & 70.22 \\
 & GPTQ & 37.22 & 51.79 & 90.43 & 72.68 & 75.19 & 65.46 \\
 & Ours & \textbf{46.11} & \textbf{60.28} & 94.94 & \textbf{77.20} & \textbf{79.02} & \textbf{71.51} \\
\midrule
\multirow{4}{*}{Qwen2-VL} & FP16 & 50.56 & 68.91 & 84.88 & 81.56 & 80.08 & 73.20 \\
\cmidrule{2-8}
 & AWQ & 45.56 & 63.18 & 82.10 & \textbf{78.44} & 77.46 & 69.35 \\
 & GPTQ & 41.89 & \textbf{67.13} & 77.79 & 72.56 & 72.22 & 66.32 \\
 & Ours & \textbf{48.00} & 66.86 & \textbf{82.30} & 77.84 & \textbf{77.82} & \textbf{70.56} \\
\bottomrule
\end{tabular}
\end{adjustbox}
\par\medskip
\centering
\caption{Per-task comparison of sensitivity-matched mixed precision, uniform CWE-20, and Ours under W3A16 and W4A8 target budgets. Bold denotes the best quantized result within each target setting. \emph{Collapse} denotes severe performance breakdown with repetitive, invalid MMMU outputs; the five-task average is therefore omitted.}
\label{tab:mixed_precision_control}
\small
\setlength{\tabcolsep}{3pt}
\begin{adjustbox}{max width=\linewidth}
\begin{tabular}{lllrrrrrr}
\toprule
Model & Target & Method & MMMU & VizWiz & ScienceQA & ChartQA & AI2D & Avg. \\
\midrule
\multirow{7}{*}{InternVL2} & FP16 & FP & 48.11 & 60.67 & 97.07 & 82.48 & 82.35 & 74.14 \\
\cmidrule{2-9}
 & \multirow{3}{*}{W3A16} & Uniform CWE-20 & 45.11 & 59.27 & 95.93 & 79.48 & \textbf{79.86} & 71.93 \\
 &  & Mixed precision & 35.11 & 54.21 & 80.07 & 55.76 & 64.67 & 57.96 \\
 &  & Ours & \textbf{47.44} & \textbf{59.77} & \textbf{96.18} & \textbf{80.12} & 79.63 & \textbf{72.63} \\
\cmidrule{2-9}
 & \multirow{3}{*}{W4A8} & Uniform CWE-20 & 44.44 & 57.25 & 96.53 & \textbf{78.88} & 78.79 & 71.18 \\
 &  & Mixed precision & Collapse & 8.50 & 30.44 & 2.80 & 21.15 & -- \\
 &  & Ours & \textbf{46.33} & \textbf{58.01} & \textbf{96.58} & 78.44 & \textbf{79.83} & \textbf{71.84} \\
\midrule
\multirow{7}{*}{LLaVA-OV} & FP16 & FP & 46.56 & 58.71 & 95.84 & 80.00 & 81.28 & 72.48 \\
\cmidrule{2-9}
 & \multirow{3}{*}{W3A16} & Uniform CWE-20 & 45.22 & \textbf{60.35} & \textbf{95.09} & \textbf{77.36} & 78.76 & 71.36 \\
 &  & Mixed precision & 36.00 & 51.87 & 87.46 & 71.12 & 72.51 & 63.79 \\
 &  & Ours & \textbf{46.11} & 60.28 & 94.94 & 77.20 & \textbf{79.02} & \textbf{71.51} \\
\cmidrule{2-9}
 & \multirow{3}{*}{W4A8} & Uniform CWE-20 & 44.78 & 55.04 & 93.41 & 74.52 & 78.01 & 69.15 \\
 &  & Mixed precision & 27.33 & 41.56 & 48.34 & 38.52 & 37.76 & 38.70 \\
 &  & Ours & \textbf{45.22} & \textbf{55.78} & \textbf{94.00} & \textbf{75.08} & \textbf{78.50} & \textbf{69.72} \\
\midrule
\multirow{7}{*}{Qwen2-VL} & FP16 & FP & 50.56 & 68.91 & 84.88 & 81.56 & 80.08 & 73.20 \\
\cmidrule{2-9}
 & \multirow{3}{*}{W3A16} & Uniform CWE-20 & 47.22 & \textbf{67.06} & 81.76 & \textbf{78.24} & \textbf{78.08} & 70.47 \\
 &  & Mixed precision & 40.00 & 55.63 & 76.05 & 71.76 & 66.35 & 61.96 \\
 &  & Ours & \textbf{48.00} & 66.86 & \textbf{82.30} & 77.84 & 77.82 & \textbf{70.56} \\
\cmidrule{2-9}
 & \multirow{3}{*}{W4A8} & Uniform CWE-20 & \textbf{47.00} & 58.87 & \textbf{80.96} & \textbf{77.28} & \textbf{77.36} & 68.29 \\
 &  & Mixed precision & 27.78 & 36.55 & 53.20 & 47.32 & 44.07 & 41.78 \\
 &  & Ours & \textbf{47.00} & \textbf{62.93} & 80.81 & 77.12 & 76.33 & \textbf{68.84} \\
\bottomrule
\end{tabular}
\end{adjustbox}
\end{table}

\subsection{LLM-backbone quantization baselines}
To compare Balanced Fitting with established language-model quantization methods, we apply AWQ~\citep{awq} and GPTQ~\citep{gptq} to the LLM backbone under W3A16 while keeping the visual encoder in FP.

Table~\ref{tab:llm_backbone_baselines} shows that Ours achieves a higher five-task average on all three models. This comparison extends the evaluation beyond LVLM-specific baselines and supports the value of allocating fitting capacity using quantization effects measured on multimodal calibration data.

\subsection{Sensitivity-matched mixed precision}
To distinguish allocating search capacity from allocating numerical precision, we construct a mixed-precision control using the same quantization effect scores. The control assigns W2/W3/W4 at an exact parameter-weighted 3-bit average for W3A16, or fixes W4 and assigns A4/A8/A16 at an exact input-width-weighted 8-bit average for W4A8, monotonically following the same QE score. Every layer uses 20-candidate CWE.

Table~\ref{tab:mixed_precision_control} compares it with uniform 20-candidate search and Ours. Ours outperforms both controls across all three models and both target budgets. Thus, directly translating the same sensitivity signal into bit-width allocation does not reproduce the benefit of Balanced Fitting in these settings. Adjusting search capacity also retains uniform inference precision; this comparison concerns the tested mixed-precision control rather than mixed-precision methods in general.

\subsection{QIG with a larger calibration set}
QIG~\citep{qig} uses 128 calibration pairs in its reported setting, whereas our main comparison fixes all methods to 64 pairs for a matched calibration-data budget. To check whether this choice accounts for our advantage, Table~\ref{tab:qig_calibration_size} additionally compares QIG with 128 pairs against Ours with 64.

Despite using half as many calibration pairs, Ours achieves the higher five-task average in five of the six model--precision settings; Qwen2-VL under W3A16 is the exception. These results support calibration-data efficiency without implying that Ours wins on every individual task.

\begin{table}[!htbp]
\centering
\caption{Five-task comparison between QIG with 128 calibration pairs and Ours with 64 pairs. Avg. is the unweighted five-task mean.}
\label{tab:qig_calibration_size}
\small
\setlength{\tabcolsep}{3pt}
\begin{adjustbox}{max width=\linewidth}
\begin{tabular}{lllrrrrrr}
\toprule
Model & Precision & Method & MMMU & VizWiz & ScienceQA & ChartQA & AI2D & Avg. \\
\midrule
\multirow{5}{*}{InternVL2} & FP16 & FP & 48.11 & 60.67 & 97.07 & 82.48 & 82.35 & 74.14 \\
\cmidrule{2-9}
 & \multirow{2}{*}{W3A16} & QIG ($n=128$) & 45.44 & 59.05 & \textbf{96.33} & 79.76 & \textbf{80.05} & 72.13 \\
 &  & Ours ($n=64$) & \textbf{47.44} & \textbf{59.77} & 96.18 & \textbf{80.12} & 79.63 & \textbf{72.63} \\
\cmidrule{2-9}
 & \multirow{2}{*}{W4A8} & QIG ($n=128$) & 45.56 & 56.19 & \textbf{96.73} & \textbf{79.32} & 79.21 & 71.40 \\
 &  & Ours ($n=64$) & \textbf{46.33} & \textbf{58.01} & 96.58 & 78.44 & \textbf{79.83} & \textbf{71.84} \\
\midrule
\multirow{5}{*}{LLaVA-OV} & FP16 & FP & 46.56 & 58.71 & 95.84 & 80.00 & 81.28 & 72.48 \\
\cmidrule{2-9}
 & \multirow{2}{*}{W3A16} & QIG ($n=128$) & 45.00 & 60.21 & \textbf{95.09} & \textbf{77.44} & \textbf{79.05} & 71.36 \\
 &  & Ours ($n=64$) & \textbf{46.11} & \textbf{60.28} & 94.94 & 77.20 & 79.02 & \textbf{71.51} \\
\cmidrule{2-9}
 & \multirow{2}{*}{W4A8} & QIG ($n=128$) & 43.22 & 53.96 & \textbf{94.25} & \textbf{75.12} & 77.88 & 68.89 \\
 &  & Ours ($n=64$) & \textbf{45.22} & \textbf{55.78} & 94.00 & 75.08 & \textbf{78.50} & \textbf{69.72} \\
\midrule
\multirow{5}{*}{Qwen2-VL} & FP16 & FP & 50.56 & 68.91 & 84.88 & 81.56 & 80.08 & 73.20 \\
\cmidrule{2-9}
 & \multirow{2}{*}{W3A16} & QIG ($n=128$) & 47.44 & \textbf{67.44} & \textbf{83.19} & \textbf{78.04} & 77.27 & \textbf{70.68} \\
 &  & Ours ($n=64$) & \textbf{48.00} & 66.86 & 82.30 & 77.84 & \textbf{77.82} & 70.56 \\
\cmidrule{2-9}
 & \multirow{2}{*}{W4A8} & QIG ($n=128$) & 45.33 & 59.78 & \textbf{81.56} & 76.68 & \textbf{76.91} & 68.05 \\
 &  & Ours ($n=64$) & \textbf{47.00} & \textbf{62.93} & 80.81 & \textbf{77.12} & 76.33 & \textbf{68.84} \\
\bottomrule
\end{tabular}
\end{adjustbox}
\end{table}

\subsection{Block-sensitivity allocation controls}
We compare two block-sensitivity controls under W4A8 using the same 64 COCO calibration pairs and CWE search setup. Both rank layers by unsigned, unweighted block reconstruction error, assigning larger grids to more sensitive layers. Fixed-grid Block independently assigns grids $\{16,18,20,22,24\}$ to five ordered sensitivity groups, placing residual layers in the middle group to retain an exact mean of 20. It does not use our grid distribution or QE-to-budget rule. Ours-grid-matched Block instead reassigns the exact grid-size multiset of Ours according to block sensitivity, preserving its total search budget. Ours uses target mean grids of 21, 18, and 19 for InternVL2, LLaVA-OV, and Qwen2-VL, respectively; the independent control therefore has a comparable, but not identical, budget.

Table~\ref{tab:block_sensitivity_controls} shows that Ours achieves the highest five-task average against both controls on all three models. Relative to Fixed-grid Block, MMMU improves on every model, by up to 2.55 points; VizWiz improves on two models, by up to 3.43 points. The controls answer complementary questions: Fixed-grid Block compares the complete allocation procedure with an independent block-based policy, whereas Ours-grid-matched Block isolates layer assignment at the same grid distribution. They are not successive ablations, and inheriting our grid distribution alone does not consistently improve block-based allocation.

\begin{table}[!htbp]
\centering
\caption{W4A8 per-task comparison with independent fixed-grid Block sensitivity. FP16 is the reference; bold denotes the best quantized result within each model.}
\label{tab:block_sensitivity_controls}
\small
\setlength{\tabcolsep}{3pt}
\begin{adjustbox}{max width=\linewidth}
\begin{tabular}{llrrrrrr}
\toprule
Model & Method & MMMU & VizWiz & ScienceQA & ChartQA & AI2D & Avg. \\
\midrule
\multirow{4}{*}{InternVL2} & FP16 & 48.11 & 60.67 & 97.07 & 82.48 & 82.35 & 74.14 \\
\cmidrule{2-8}
 & Fixed-grid Block & 43.78 & 55.93 & \textbf{96.63} & 78.48 & 78.89 & 70.74 \\
 & Ours-grid-matched Block & 44.00 & 57.47 & 96.33 & \textbf{78.80} & 79.57 & 71.23 \\
 & Ours & \textbf{46.33} & \textbf{58.01} & 96.58 & 78.44 & \textbf{79.83} & \textbf{71.84} \\
\midrule
\multirow{4}{*}{LLaVA-OV} & FP16 & 46.56 & 58.71 & 95.84 & 80.00 & 81.28 & 72.48 \\
\cmidrule{2-8}
 & Fixed-grid Block & 44.22 & \textbf{56.19} & \textbf{94.79} & 74.40 & \textbf{78.50} & 69.62 \\
 & Ours-grid-matched Block & \textbf{45.22} & 54.56 & 94.40 & 74.64 & 78.04 & 69.37 \\
 & Ours & \textbf{45.22} & 55.78 & 94.00 & \textbf{75.08} & \textbf{78.50} & \textbf{69.72} \\
\midrule
\multirow{4}{*}{Qwen2-VL} & FP16 & 50.56 & 68.91 & 84.88 & 81.56 & 80.08 & 73.20 \\
\cmidrule{2-8}
 & Fixed-grid Block & 46.44 & 59.50 & \textbf{80.81} & 76.76 & \textbf{77.33} & 68.17 \\
 & Ours-grid-matched Block & 45.56 & 60.24 & 80.52 & 76.68 & 76.68 & 67.94 \\
 & Ours & \textbf{47.00} & \textbf{62.93} & \textbf{80.81} & \textbf{77.12} & 76.33 & \textbf{68.84} \\
\bottomrule
\end{tabular}
\end{adjustbox}
\end{table}

\subsection{Computational cost}
Table~\ref{tab:parallel_calibration_time} reports W4A8 calibration wall-clock time on COCO-64, excluding model loading and evaluation. Ours runs three quantization effect analyses concurrently on three NVIDIA RTX A6000 GPUs (48GB each); each CWE search run and both baselines use one GPU. Thus, this comparison measures wall-clock time under different GPU allocations, rather than matched compute.

Ours has the shortest CWE search time on all three models, which offsets much of the additional effect-analysis time. Its total elapsed time is lower than QIG on all three models and lower than MBQ on LLaVA-OV, while MBQ remains faster on InternVL2 and Qwen2-VL. The benefit is therefore comparable calibration turnaround under the stated hardware allocation, not a claim of uniformly lower computational cost. Calibration is performed once and introduces no inference overhead.

\begin{table}[!htbp]
\centering
\caption{W4A8 calibration time on NVIDIA RTX A6000 48GB GPUs using 64 COCO samples; model loading and evaluation are excluded. Times are wall-clock times.}
\label{tab:parallel_calibration_time}
\small
\setlength{\tabcolsep}{3pt}
\begin{adjustbox}{max width=\linewidth}
\begin{tabular}{llrrr}
\toprule
Model & Method & Analysis & CWE search & Total \\
\midrule
\multirow{3}{*}{InternVL2} & MBQ & 0m 56s & 24m 12s & 25m 08s \\
 & QIG & 4m 24s & 23m 30s & 27m 54s \\
 & Ours & 21m 02s & 6m 28s & 27m 30s \\
\midrule
\multirow{3}{*}{LLaVA-OV} & MBQ & 1m 10s & 35m 02s & 36m 12s \\
 & QIG & 7m 04s & 35m 31s & 42m 35s \\
 & Ours & 32m 11s & 1m 55s & 34m 06s \\
\midrule
\multirow{3}{*}{Qwen2-VL} & MBQ & 1m 01s & 26m 21s & 27m 22s \\
 & QIG & 4m 49s & 26m 57s & 31m 46s \\
 & Ours & 18m 59s & 8m 41s & 27m 40s \\
\bottomrule
\end{tabular}
\end{adjustbox}
\end{table}

\clearpage
\section{Quantization Effect and Prediction Analyses}
\label{sec:quantization_prediction_analysis}

\subsection{Stability metric definitions}
To test whether the allocation signal depends on a particular calibration subset, Table~\ref{tab:qe_subset_stability} compares three disjoint calibration subsets with an independent held-out subset. Components are ranked by calibration mean relative absolute QE. QE mass is the captured fraction of absolute QE. Three-subset agreement requires matching signs across all calibration subsets; held-out agreement compares their majority sign with the held-out sign. Beneficial held-out agreement restricts this comparison to calibration-beneficial effects. Ranking uses calibration data only, so the held-out subset tests whether the selected effects retain their direction on unseen samples.

Agreement is strongest for high-magnitude effects, including beneficial ones, supporting both refinement and coarsening decisions. This analysis tests the stability of the allocation signal, rather than the variance of final downstream performance across calibration seeds.

\subsection{Controlled search granularity}
To examine whether coarser fitting can help independently of bit-width allocation, we use W4A8 and COCO-64 and vary only the CWE grids that Ours coarsens below the 20-point reference, keeping precision and all remaining grids fixed. Fine replaces these counts with 25 for InternVL2 and 23 for LLaVA-OV and Qwen2-VL; Mid uses 20, and Coarse retains the original Ours counts. All counts uniformly discretize the same exponent interval $[0,1]$, so their interior candidates need not be nested. After recalibration, we compare aligned per-channel scale vectors.

Table~\ref{tab:granularity_scale_similarity} shows that Coarse achieves the highest five-task average on all three models despite highly similar Fine--Coarse scale vectors. The intermediate setting is not consistently better than Fine, so the result does not suggest a monotonic benefit from reducing candidates. Instead, it supports selectively restricting fitting capacity according to the QE-guided allocation.

\begin{table}[!htbp]
\centering
\caption{W4A8 per-task results under controlled search granularity. Fine--Coarse scale-vector cosine similarity is 0.9895 / 0.9955 / 0.9902 for InternVL2 / LLaVA-OV / Qwen2-VL, respectively. Bold denotes the best quantized result.}
\label{tab:granularity_scale_similarity}
\small
\setlength{\tabcolsep}{3pt}
\begin{adjustbox}{max width=\linewidth}
\begin{tabular}{llrrrrrr}
\toprule
Model & Setting & MMMU & VizWiz & ScienceQA & ChartQA & AI2D & Avg. \\
\midrule
\multirow{4}{*}{InternVL2} & FP16 & 48.11 & 60.67 & 97.07 & 82.48 & 82.35 & 74.14 \\
\cmidrule{2-8}
 & Fine & 45.11 & 57.39 & \textbf{96.63} & \textbf{78.96} & 79.15 & 71.45 \\
 & Mid & 45.22 & 57.26 & \textbf{96.63} & 78.52 & 78.14 & 71.15 \\
 & Coarse (Ours) & \textbf{46.33} & \textbf{58.01} & 96.58 & 78.44 & \textbf{79.83} & \textbf{71.84} \\
\midrule
\multirow{4}{*}{LLaVA-OV} & FP16 & 46.56 & 58.71 & 95.84 & 80.00 & 81.28 & 72.48 \\
\cmidrule{2-8}
 & Fine & 43.67 & 55.36 & \textbf{94.35} & \textbf{75.08} & 78.11 & 69.31 \\
 & Mid & 42.22 & \textbf{55.93} & 93.51 & 75.04 & 78.24 & 68.99 \\
 & Coarse (Ours) & \textbf{45.22} & 55.78 & 94.00 & \textbf{75.08} & \textbf{78.50} & \textbf{69.72} \\
\midrule
\multirow{4}{*}{Qwen2-VL} & FP16 & 50.56 & 68.91 & 84.88 & 81.56 & 80.08 & 73.20 \\
\cmidrule{2-8}
 & Fine & 46.89 & 61.58 & 80.42 & 77.00 & 76.59 & 68.50 \\
 & Mid & 46.89 & 61.64 & 80.61 & 76.52 & \textbf{77.20} & 68.57 \\
 & Coarse (Ours) & \textbf{47.00} & \textbf{62.93} & \textbf{80.81} & \textbf{77.12} & 76.33 & \textbf{68.84} \\
\bottomrule
\end{tabular}
\end{adjustbox}
\end{table}

\subsection{Full-set FP-answer transitions}
To determine whether the gains primarily come from correcting FP mistakes or avoiding new errors, we evaluate all paired FP--PTQ predictions on MMMU (900), VizWiz (4,319), ScienceQA (2,017), ChartQA (2,500), and AI2D (3,088): 12,824 examples per model and precision. Correctness follows each benchmark's evaluator, with positive VQA credit counted as correct for VizWiz. Transition rates use their corresponding FP-wrong or FP-correct subset; fixes/new errors is a ratio of counts.

Table~\ref{tab:fp_transitions} shows that Ours consistently reduces new errors and improves retention and the fixes/new-errors ratio, while its correction rate is not uniformly highest. The aggregate gains therefore primarily reflect better preservation of correct FP behavior and fewer newly introduced errors, complemented by corrections of some FP mistakes. This full-set analysis provides context for the qualitative examples rather than relying on selected corrections alone.

\begin{table}[!htbp]
\centering
\caption{Full-set FP-answer transitions under W3A16 and W4A8. Rates are conditional on the corresponding FP-correct or FP-wrong subset.}
\label{tab:fp_transitions}
\small
\setlength{\tabcolsep}{3pt}
\begin{adjustbox}{max width=\linewidth}
\begin{tabular}{lllrrrr}
\toprule
Model & Precision & Method & \shortstack{FP wrong $\to$\\correct (\%)} & \shortstack{FP correct $\to$\\wrong (\%)} & \shortstack{FP correct $\to$\\correct (\%)} & \shortstack{Fixes/\\new errors} \\
\midrule
\multirow{6}{*}{InternVL2} & \multirow{3}{*}{W3A16} & MBQ & \textbf{15.92} & 6.99 & 93.01 & 0.63 \\
 &  & QIG & 14.27 & 6.70 & 93.30 & 0.59 \\
 &  & Ours & 15.38 & \textbf{6.23} & \textbf{93.77} & \textbf{0.68} \\
\cmidrule{2-7}
 & \multirow{3}{*}{W4A8} & MBQ & \textbf{19.13} & 9.00 & 91.00 & 0.59 \\
 &  & QIG & 16.53 & 8.98 & 91.02 & 0.51 \\
 &  & Ours & 19.02 & \textbf{8.32} & \textbf{91.68} & \textbf{0.63} \\
\midrule
\multirow{6}{*}{LLaVA-OV} & \multirow{3}{*}{W3A16} & MBQ & 22.52 & 8.59 & 91.41 & 0.80 \\
 &  & QIG & 21.78 & 8.33 & 91.67 & 0.80 \\
 &  & Ours & \textbf{23.05} & \textbf{7.94} & \textbf{92.06} & \textbf{0.88} \\
\cmidrule{2-7}
 & \multirow{3}{*}{W4A8} & MBQ & \textbf{23.02} & 12.81 & 87.19 & 0.55 \\
 &  & QIG & 20.01 & 12.26 & 87.74 & 0.50 \\
 &  & Ours & 22.19 & \textbf{10.70} & \textbf{89.30} & \textbf{0.63} \\
\midrule
\multirow{6}{*}{Qwen2-VL} & \multirow{3}{*}{W3A16} & MBQ & 20.28 & 9.74 & 90.26 & 0.55 \\
 &  & QIG & 19.76 & 9.50 & 90.50 & 0.55 \\
 &  & Ours & \textbf{20.73} & \textbf{8.70} & \textbf{91.30} & \textbf{0.63} \\
\cmidrule{2-7}
 & \multirow{3}{*}{W4A8} & MBQ & 19.99 & 13.44 & 86.56 & 0.40 \\
 &  & QIG & \textbf{20.58} & 12.22 & 87.78 & 0.45 \\
 &  & Ours & 19.80 & \textbf{11.02} & \textbf{88.98} & \textbf{0.48} \\
\bottomrule
\end{tabular}
\end{adjustbox}
\end{table}

\subsection{QE under joint quantization}
Isolated component effects may change when multiple components are quantized together. To examine whether their direction remains informative in that setting, we compare isolated QE with restoration-based QE obtained by restoring one layer-component to FP in an otherwise quantized model. We define restoration-based QE as $\mathcal{L}_{\mathrm{restore}}-\mathcal{L}_{\mathrm{joint}}$, the calibration loss after restoration minus that of the jointly quantized model; positive values indicate that retaining the quantized component lowers loss. Both use COCO-64 with allocation and CWE search disabled. QE mass uses absolute isolated QE; beneficial agreement considers only beneficial isolated effects.

Table~\ref{tab:joint_qe_agreement} reports sign agreement between isolated and restoration-based QE, with stronger agreement for high-magnitude effects, including beneficial effects. This supports using isolated measurements to identify influential components even in jointly quantized models. The imperfect agreement also indicates that isolated effects do not fully capture interactions among quantized components.

\begin{table}[!htbp]
\centering
\caption{Agreement between isolated QE and restoration-based QE in jointly quantized W3A16 and W4A8 models. Parentheses give agreement counts.}
\label{tab:joint_qe_agreement}
\small
\setlength{\tabcolsep}{3pt}
\begin{adjustbox}{max width=\linewidth}
\begin{tabular}{lcccc}
\toprule
QE-ranked subset & Count & QE mass & Joint sign agreement & Beneficial joint agreement \\
\midrule
Top 50\% & 176 & 91.50\% & 68.20\% (120/176) & 75.60\% (34/45) \\
Top 25\% & 88 & 75.80\% & 81.80\% (72/88) & 95.50\% (21/22) \\
Top 10\% & 38 & 58.20\% & \textbf{89.50\% (34/38)} & \textbf{100\% (11/11)} \\
\bottomrule
\end{tabular}
\end{adjustbox}
\end{table}

\clearpage
\section{Additional Qualitative Results}

\begin{figure}
    \centering
    \includegraphics[width=\linewidth]{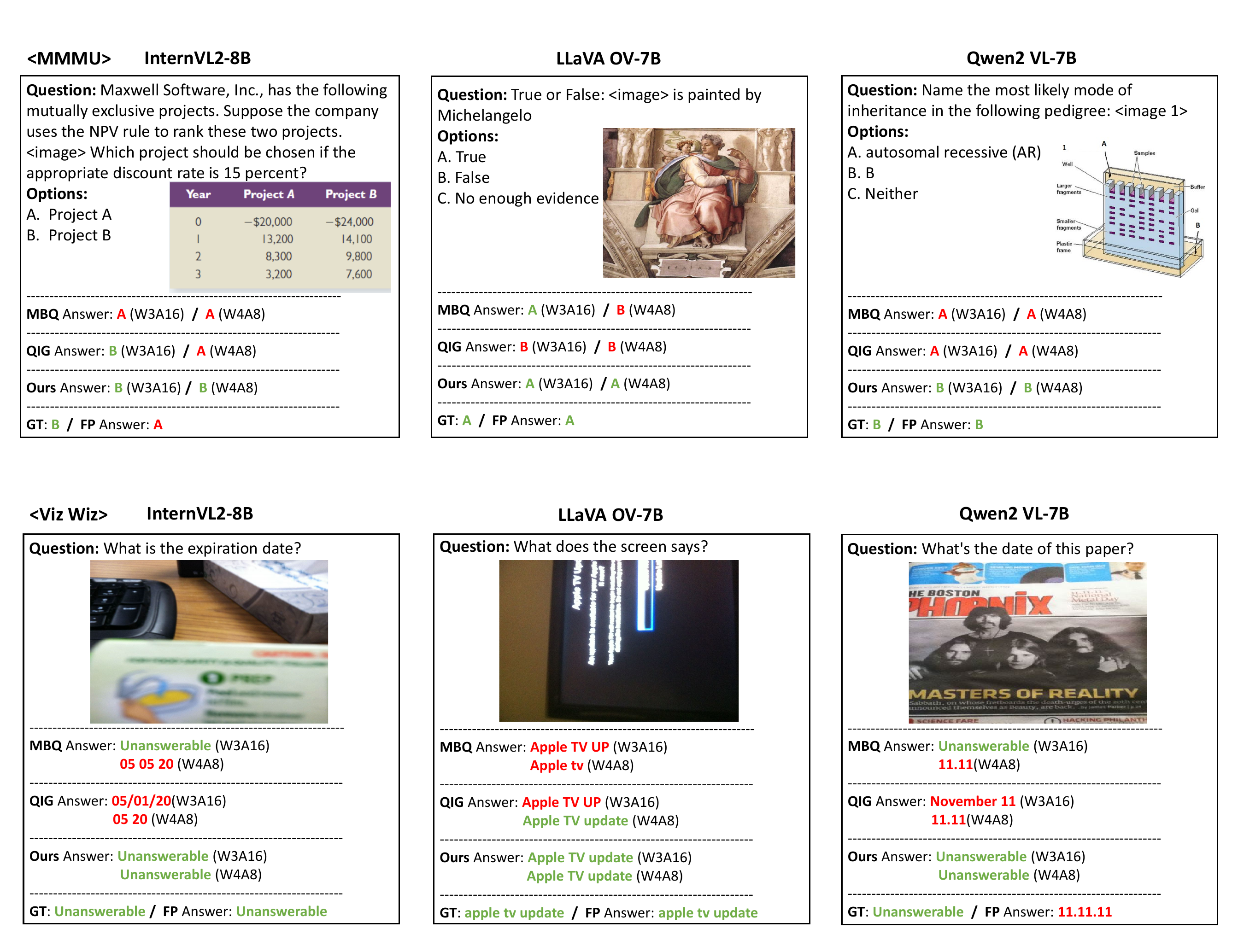}
    \caption{Additional qualitative comparison on MMMU and VizWiz.}
    \label{fig:qualitative_results_appendix1}
\end{figure}

\begin{figure}[t]
    \centering
    \includegraphics[width=\linewidth]{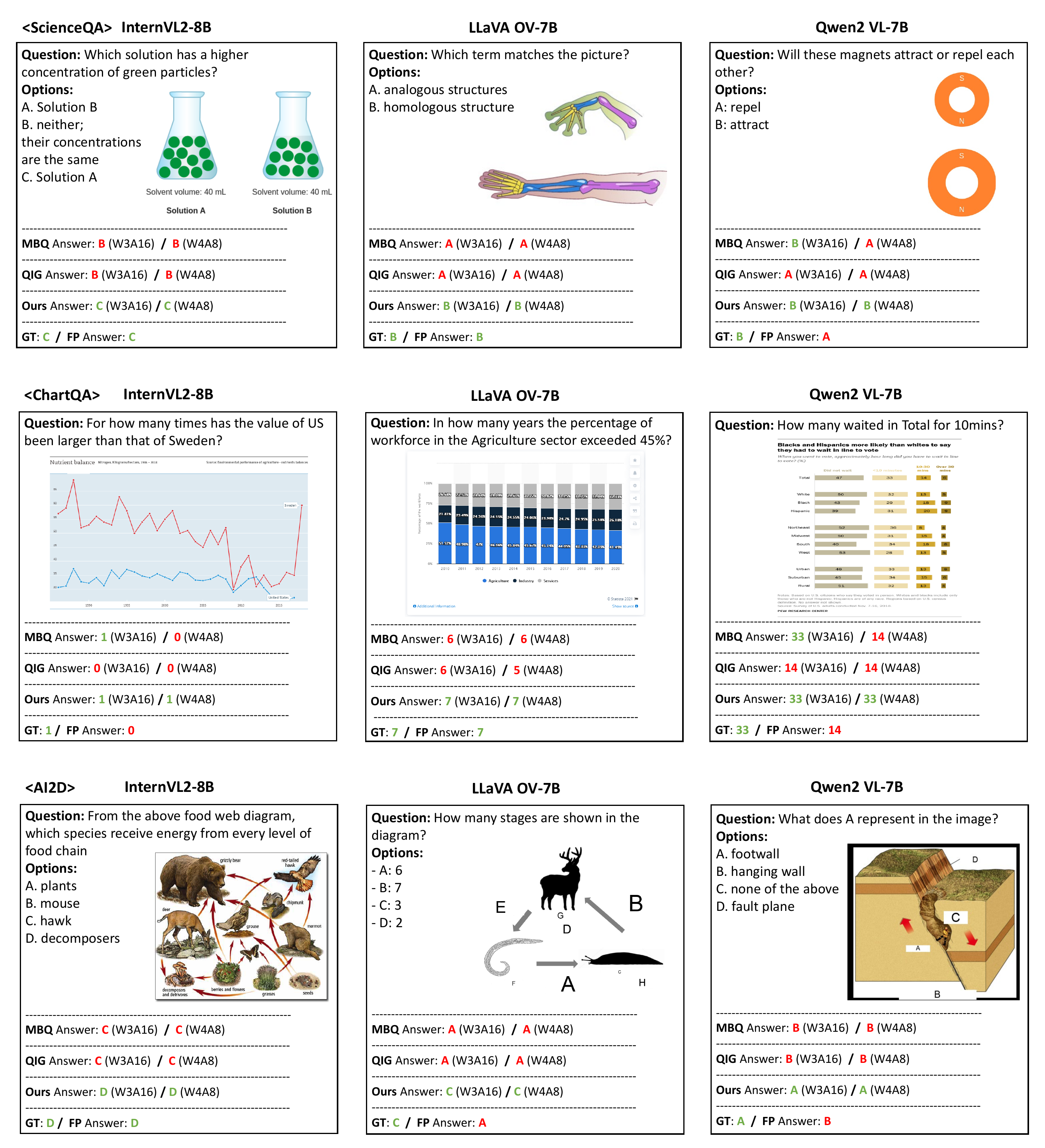}
    \caption{Additional qualitative comparison on ScienceQA, ChartQA, and AI2D.}
    \label{fig:qualitative_results_appendix2}
\end{figure}

Figure~\ref{fig:qualitative_results_appendix1} and Figure~\ref{fig:qualitative_results_appendix2} provide additional qualitative comparisons on five vision-language benchmarks~\citep{mmmu,vizwiz,scienceqa,chartqa,ai2d} across three LVLMs~\citep{internvl,llavaov,qwen2}.

Consistent with the full-set transition analysis, baseline PTQ methods (MBQ~\citep{mbq} and QIG~\citep{qig}) can replicate incorrect FP predictions or introduce specific but erroneous outputs under low-bit settings, particularly in W4A8. In contrast, our method more reliably preserves correct FP predictions and introduces fewer new errors; selected examples also show corrections of FP errors. 
On MMMU~\citep{mmmu}, the examples illustrate that selective coarse fitting need not reproduce an incorrect FP answer and can recover the correct choice. 
On VizWiz~\citep{vizwiz}, where inputs are often ambiguous or low-quality, baseline methods frequently generate incorrect specific answers, while our method produces more reliable outputs, including correctly identifying unanswerable cases. Figure~\ref{fig:qualitative_results_appendix2} shows similar trends on ScienceQA~\citep{scienceqa}, ChartQA~\citep{chartqa}, and AI2D~\citep{ai2d}. 
Across these structured reasoning and diagram understanding tasks, our method primarily benefits from preserving valid FP behavior while using regularization-like constrained fitting to reduce calibration-specific overfitting.

Overall, these results suggest that our approach better balances precise fitting and regularization-like capacity control, preserving valid FP behavior and improving generalization under aggressive quantization.

\clearpage
\end{document}